\documentclass[11pt,a4paper,onecolumn,english,english,times]{article}
\usepackage[letterpaper]{geometry}
\usepackage[T1]{fontenc}
\usepackage[latin9]{inputenc}
\usepackage{color}
\usepackage{array}
\usepackage{float}
\usepackage{amsmath}
\usepackage{graphicx}
\usepackage{amssymb}

\makeatletter

\providecommand{\tabularnewline}{\\}
\floatstyle{ruled}
\newfloat{algorithm}{tbp}{loa}
\floatname{algorithm}{Algorithm}

\@ifundefined{definecolor}
 {\usepackage{color}}{}

\usepackage{array}
\usepackage{float}

\makeatletter

\providecommand{\tabularnewline}{\\}
\floatstyle{ruled}
\newfloat{algorithm}{tbp}{loa}
\floatname{algorithm}{Algorithm}

\@ifundefined{definecolor}
 {\@ifundefined{definecolor}
 {\usepackage{color}}{}
}{}

\usepackage{array}
\usepackage{float}
\usepackage{amstext}

\usepackage{amsthm}

\makeatletter

\providecommand{\tabularnewline}{\\}
\floatstyle{ruled}
\newfloat{algorithm}{tbp}{loa}
\floatname{algorithm}{Algorithm}

\@ifundefined{definecolor}
 {\@ifundefined{definecolor}
 {\@ifundefined{definecolor}
 {\usepackage{color}}{}
}{}
}{}
\@ifundefined{definecolor}
 {\@ifundefined{definecolor}
 {\@ifundefined{definecolor}
 {\@ifundefined{definecolor}
 {\usepackage{color}}{}
}{}
}{}
}{}

\usepackage{array}
\usepackage{float}

\usepackage{times}

\usepackage{subfigure}

\floatstyle{ruled}
\newfloat{algorithm}{tbp}{loa}
\floatname{algorithm}{Algorithm}

\usepackage{algorithmic}
\usepackage{algorithm}

\font\elvbf  = cmbx10 scaled 1100

\def\section{\@startsection {section}{1}{\z@}
   {14pt plus 2pt minus 2pt}{14pt plus 2pt minus 2pt} {\large\bf}} 
\def\subsection{\@startsection {subsection}{2}{\z@}
   {13pt plus 2pt minus 2pt}{13pt plus 2pt minus 2pt} {\elvbf}}

\newcommand{\Section}[1]{\section{\hskip -1em.~#1}} 
\newcommand{\SubSection}[1]{\subsection{\hskip -1em.~#1}}

\providecommand{\tabularnewline}{\\}

\long\def\narenc#1{[{\color{red}todo: Naren writes: \it #1}]}

\newtheorem{lemma}{Lemma}
\newtheorem{theorem}{Theorem}

\newtheorem{problem}{Problem}

\theoremstyle{definition}
\newtheorem{definition}{Definition}

\newcommand{\comment}[1]{}

\makeatother
\usepackage{babel}

\makeatother

\usepackage{babel}

\makeatother

\usepackage{babel}

\let\oldenumerate=\enumerate
\def\enumerate{\oldenumerate%
\setlength{\itemsep}{0pt}\setlength{\parsep}{0pt}}%

\makeatother

\usepackage{babel}

\makeatother

\usepackage{babel}

\begin{document}

\title{Efficiently Discovering Hammock Paths from Induced Similarity Networks}

\author{M. Shahriar Hossain, Michael Narayan, Naren Ramakrishnan\\
\it Department of Computer Science, Virginia Tech, Blacksburg, VA 24061\\
\it Email: \{msh, mnarayan, naren\}@vt.edu}

\date{}

\maketitle

\thispagestyle{empty} 
\begin{abstract}
\noindent Similarity networks are important abstractions in many information
management applications such as recommender systems, corpora analysis,
and medical informatics. For instance, in a recommender system, by
inducing similarity networks between movies rated similarly by users,
we can aim to find the global structure of connectivities underlying
the data, and use the network to posit connections between given entities.
We present an algorithmic framework to efficiently find paths in an
induced similarity network \textit{without materializing the network
in its entirety}. Our framework introduces the notion of `hammock'
paths which are generalizations of traditional paths in bipartite
graphs. 
Given starting and ending objects of interest, it explores candidate
objects for path following, and heuristics to admissibly estimate
the potential for paths to lead to a desired destination. We present
three diverse applications, modeled after the Netflix dataset, a broad
subset of the PubMed corpus, and a database of clinical trials. Experimental
results demonstrate the potential of our approach for unstructured
knowledge discovery in similarity networks. 
\end{abstract}
\vspace{-0.2in}
 \Section{Introduction} \vspace{-0.15in}

There is significant current interest in modeling and understanding
network structures, especially in online social communities, web graphs,
and biological networks. We focus here on (unipartite) similarity
networks induced from bipartite graphs, such as a movies $\times$
people dataset. Two movies can be connected if they have been rated
similarly by sufficient number of people; this is the basis for the
popular item-based recommendation algorithms~\cite{karypis01}. A
similarity network thus exposes an indirect level of clustering, community
formation, and organization that is not immediately apparent from
the bipartite network.

Similarity networks are key abstractions in recommender systems but
they also find uses in other information management applications such
as collection analysis and medical informatics. We focus on how these
networks can be used for \textit{exploratory discovery}, i.e., to
see how similarities can be composed to reach potentially distant
entities. For instance, how is the movie `Roman Holiday' (a romantic
classic) connected to `Terminator 3' (a Sci-Fi movie)? What are the
in-between movies and waypoints that help link these two disparate
movies? Are these waypoints characteristic to the domain or to the
dataset? For recommender systems, these questions are especially germane
to tasks such as serendipitous recommendation, gift buying, and characterizing
the movie watching interests of its users.

In addition to recommender systems, we consider similarity networks
induced from literature and semi-structured sources of data. Here
path finding has applications to literature-based discovery (similar
to the ARROWSMITH program~\cite{Smalheiser_ARROWSMITH}) and clinical
diagnosis. For instance, how is congestive heart failure related to
kidney disease? The discovered waypoints correspond to possibilities
by which different disease response pathways interact with each other.

Admittedly, these questions can be answered by first inducing the
similarity network and running shortest path queries over it, but
we desire to find paths \textit{without materializing the entire network}.
This would be wasteful because only a subnetwork is likely to be traversed
in response to a query. Instead, we seek to pre-process the original
bipartite graph into suitable data structures that can then be harnessed
to find chains of connections.

In this paper, we present an algorithmic framework for traversing
paths using the notion of `hammock' paths which are generalization
of traditional paths. Our framework is exploratory in nature so that,
given starting and ending objects of interest, it explores candidate
objects for path following, and heuristics to admissibly estimate
the potential for paths to lead to a desired destination. Our key
contributions are:

\begin{figure*}[t]
 \centering

\includegraphics[scale=0.5]{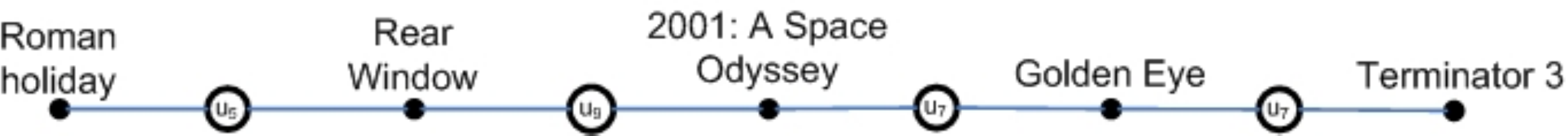}

\includegraphics[scale=0.45]{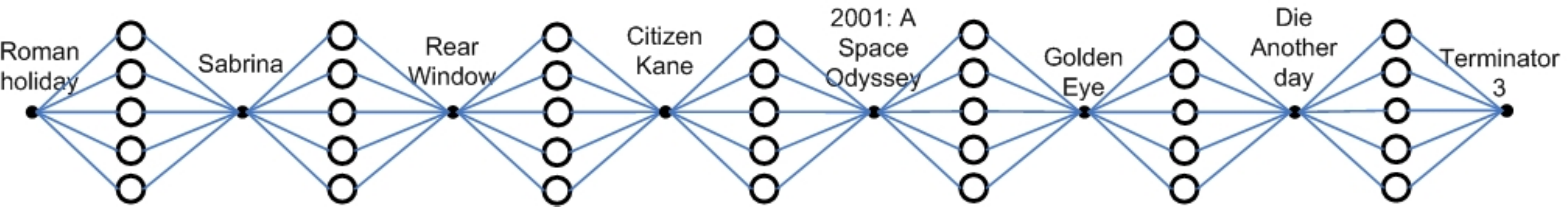}

\includegraphics[scale=0.45]{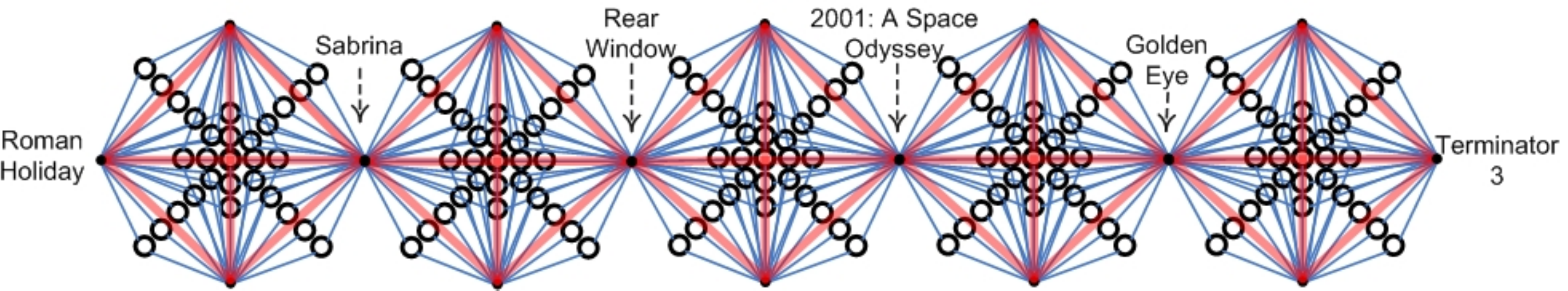}

\caption{\small{(\emph{Top}) A simple path (hammock width=1) beginning at romantic
classic {}``Roman Holiday'' and ending at the Sci-Fi \& Fantasy
movie {}``Terminator III''. (\emph{Middle}) A hammock path between
the same entities but with hammock width=5. (\emph{Bottom}) A hammock
path between the same entities with an additional clique size requirement
of 4. Observe that each hammock inside the cliques continues to have
width 5. Notice that the path length increases from (\emph{Top}) to
(\emph{Middle}) and then drops in (\emph{Bottom}).} \label{fig:Movie-recommendationsExample}}


\end{figure*}

\itemsep=0pt
\begin{enumerate}
\item We formulate the path finding problem over similarity networks in
terms of hammocks and cliques, two intuitive-to-understand constructs
for navigating similarity networks. Hammocks are ways to impose tighter
requirements over individual links (leading to longer paths) and cliques
are ways to coalesce multiple links (resulting in shorter paths).

\item We present an algorithmic framework that finds hammock paths in both
binary datasets (bipartite graphs) and vector-valued datasets (weighted
bipartite graphs). Our framework uses a concept lattice representation
as an in-memory data structure to organize the search for paths and
introduces admissible heuristics that quickly prune out unpromising
paths. 
\item We present compelling experimental results on three diverse datasets:
the Netflix dataset of movie ratings, a significant portion of the
PubMed corpus, and a database of clinical trials. Experimental results
demonstrate the scalability and accuracy of our approach as well as
its potential for unstructured knowledge discovery. 
\end{enumerate}

\vspace{-0.2in}
 \Section{Background} \vspace{-0.15in}

The formulations we study can be intuitively understood using Fig.~\ref{fig:Movie-recommendationsExample}.
A simple path between movies can be induced through co-raters, as
shown in Fig.~\ref{fig:Movie-recommendationsExample} (top): `Roman
Holiday' was rated by user $u_{5}$ who also rated `Rear Window' which
was also rated by $u_{9}$, and so on. In practice, paths are induced
between two movies only if a sufficient number of people have rated
them and rated them similarly. This leads to a sequence of `hammocks'
as shown in Fig.~\ref{fig:Movie-recommendationsExample} (middle).
A final level of generalization is to organize the hammocks into groups
so that we traverse cliques of movies (with a hammock between every
pair of them), see Fig.~\ref{fig:Movie-recommendationsExample} (bottom).
By finding such hammocks and traversing them systematically, similarities
can `diffuse' to reach possibly distant entities. 
Note that the path length increases as the hammock requirement is
strengthened and then decreases as the clique requirement is imposed.
Work closest to ours can be found in the graph modeling, redescription
mining, and lattice-based information retrieval literature.

\noindent \textbf{Graph modeling:} The use of the word `hammock' for
induced similarity networks appears to have been first made in~\cite{Mirza_Studying}
although this work does not aim to find paths. 
\comment{\narenc{Shahriar, remove the next sentence, but sprinkle
citations to these guys later on in the paper.}} 
The notion of kNC-plots was introduced in~\cite{Kumar_Connectivity}
where $k$ denotes what we refer to as hammock width in this paper.
Rather than finding local paths, this paper is focused on finding
the global connectivity structure of the induced networks as the hammock
width is increased. It is also restricted to binary spaces whereas
we focus on vector valued spaces as well. 
Random intersection graphs~\cite{Stark_The} are a class of theoretical
models proposed in the random graph community. These models randomly
assign each vertex with a subset of a given set and posit edges if
these subsets intersect. Under these modeling assumptions, connectivity
and other properties of these graphs can be characterized \cite{Eppstein_Sparsification}.

\noindent \textbf{Redescription mining:} Redescriptions, a pattern
class introduced in~\cite{Ramakrishnan_Turning}, induce subsets
of data that share strong overlap. A hammock in our notation can be
viewed as a redescription between the objects it connects. Kumar et
al.~\cite{Kumar_Algorithms} study the problem of `storytelling'
over a space of descriptors, which is the goal of finding a sequence
(story) of redescriptions between two subsets by positing intermediate
subsets that share sufficient overlap with their neighbors in the
story. Although this is similar to our objective, the heuristic innovations
in~\cite{Kumar_Algorithms} are restricted to binary spaces as well
and cannot find paths of the same expressiveness (hammocks organized
into cliques) and with the same efficiencies as done here.

\noindent \textbf{Lattice based information retrieval:} The use of
concept lattices as organizing data structures for fast retrieval,
query, and data mining is not new~\cite{Narayanaswamy_A,Pedersen_A}.
Our work is different in both the theoretical framework by which the
lattice concepts are used to build chains and in their uses/applications.

\vspace{-0.1in}
 \Section{Problem Formulation} \vspace{-0.15in}

We begin by formally defining the space over which similarity networks
are induced.

\begin{definition} \label{Def_database}A dataset ${\cal D}=({\cal O},{\cal F},{\cal V})$
is a set of objects ${\cal O}$, a set of features ${\cal F}$, and
a relation ${\cal V}\subseteq{\cal O}\times{\cal F}$. \end{definition}

\begin{figure}
\centering \includegraphics[scale=0.35]{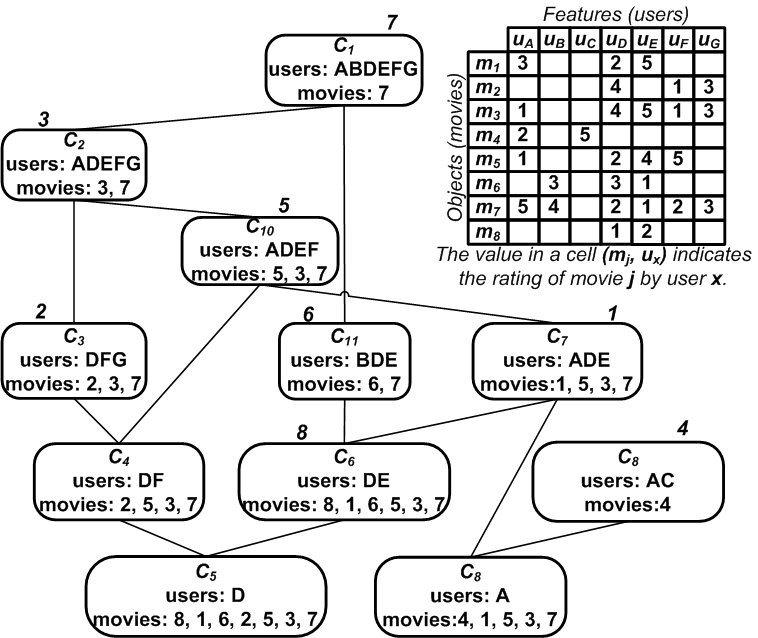}

\vspace{-0.05in}

\caption{\small{A dataset and its concept lattice.} \comment{ \narenc{The table
has to be flipped. Make the rows to be movies and the columns to be
features. Give a label for the vertical dimension of the table as
`Objects (movies).' Give a label for the horizontal dimension of the
table as `Features (users).' Also the sentence at the bottom of the
table is wrong. It is using capital X when it should use small x.}}}

\label{fig:A-concept-lattice_minsup=00003D00003D00003D00003D00003D00003D00003D00003D00003D2}

\vspace{-0.2in}

\end{figure}

We can thus think of each $f\in{\cal F}$ as inducing a subset of
${\cal O}$, which we call $o(f)$. We further require that the objects
induced by ${\cal F}$ form a covering of ${\cal O}$, i.e. ${\displaystyle \bigcup_{f\in{\cal F}}o(f)={\cal O}}$.
In the applications studied here, the $({\cal O},{\cal F})$ are (movies,
users), (documents, terms), and (clinical trials, keywords), respectively,
with the obvious meaning of ${\cal V}$ in each case. In Fig.~\ref{fig:A-concept-lattice_minsup=00003D00003D00003D00003D00003D00003D00003D00003D00003D2}
we introduce a dataset which we will use as a running example. Here
the objects are movies (${\cal O}=\{m_{1},m_{2},m_{3},m_{4},m_{5},m_{6},m_{7},m_{8}\}$)
and the features are users (${\cal F}=\{u_{A},u_{B},u_{C},u_{D},u_{E},u_{F},u_{G}\}$).

A note on notation: We use calligraphic ${\cal O}$ and ${\cal F}$
to represent the complete object and feature sets for some database.
Lower case letters $o$ and $f$ are used to represent individual
members of ${\cal O}$ and ${\cal F}$, respectively. To represent
subsets of ${\cal O}$ and ${\cal F}$ we use upper case letters $O$
and $F$, resp.

For a feature set $F\subseteq{\cal F}$, we overload notation and
define the operator $o(F):{\cal F}\rightarrow{\cal O}$ as the set
of objects which are associated with all of the features in $F$,
i.e.,

\vspace{-0.1in}
 \[
o(F)={\displaystyle \bigcap_{f\in F}o(f)}\]

\vspace{-0.05in}
 We also define a parallel operator $f(O)$ as the set of features
associated with the object set $O$. 
In Fig.~\ref{fig:A-concept-lattice_minsup=00003D00003D00003D00003D00003D00003D00003D00003D00003D2},
we have $o(\{u_{A},u_{D},u_{E}\})=\{m_{1},m_{3},m_{5},m_{7}\}$ and
$f(\{m_{1},m_{3},m_{5},m_{7}\})=\{u_{A},u_{D},u_{E}\}$.

We can now define the closure operator $c:2^{{\cal F}}\rightarrow2^{{\cal F}}$
as $c(F)=f(o(F))$. A feature set $F$ is closed if and only if $c(F)=F$.
(A parallel closure operator exists on $2^{{\cal O}}$, but is not
necessary for our purposes.) Using our running example from Figure~\ref{fig:A-concept-lattice_minsup=00003D00003D00003D00003D00003D00003D00003D00003D00003D2}
we see that feature set $\{u_{A},u_{D},u_{E}\}$ is closed, whereas
$\{u_{A},u_{D}\}$ is not, since $c(\{u_{A},u_{D}\})=\{u_{A},u_{D},u_{E}\}\neq\{u_{A},u_{D}\}$.

\begin{definition} For each feature $f\in{\cal F}$, we define a
predicate $p_{f}(o)$ such that $p_{f}(o)$ is true if and only if
$o\in o(f)$. The set of all such predicates is denoted by ${\cal P}$.
A \textit{descriptor} is a boolean expression over ${\cal P}$. \end{definition}

A descriptor thus provides a mechanism to specify a set of objects
which correspond to some Boolean expression on features. Again referring
to Figure~\ref{fig:A-concept-lattice_minsup=00003D00003D00003D00003D00003D00003D00003D00003D00003D2}
we might define a descriptor $u_{A}(o)\vee u_{B}(o)$, which would
correspond to all the movies which were rated by user A, by user B,
or by both; it induces the set $\{m_{1},m_{3},m_{5},m_{6},m_{7}\}$.
It is easy to see that we can create a descriptor for the objects
in $o(F)$ by creating a conjunction over the features in $F$. For
simplicity, we often speak of descriptor $F$ for some feature set,
which refers to this conjunction.

We employ the notion of \textit{redescriptions} introduced by Ramakrishnan
et al.~\cite{Ramakrishnan_Turning} to capture similarities between
descriptors. If for descriptors $F$ and $F'$, where $F$ is tautologically
distinct from $F'$ (i.e. $F\vee\neg F'$ is not a tautology), it
is the case that $o(F)=o(F')$, we say that $F$ and $F'$ are \textit{redescriptions}
of each other, as they induce the same set of objects. In other words,
a redescription provides (at least) two different ways to describe
a given object set. We next introduce the concept of a redescription
set:

\begin{definition} A \textit{redescription set} for a descriptor
$F$, $R_{F}=(O',F')$ is a tuple where $O'=o(F)$ and $F'=c(F)$.
\end{definition}

Fig.~\ref{fig:A-concept-lattice_minsup=00003D00003D00003D00003D00003D00003D00003D00003D00003D2}
shows many redescription sets organized alongside a concept lattice
(to be defined soon). One redescription set is: ${\bf C}_{7}=(\{m_{1},m_{3},m_{5},m_{7}\},\{u_{A},u_{D},u_{E}\})$,
which would correspond to both descriptors $\{u_{A},u_{D},u_{E}\}$
and $\{u_{A},u_{D}\}$.

We say that a descriptor $F$ is a relaxation of descriptor $F'$,
denoted by ($F\leq F')$, if the feature set $F=\{f_{1},f_{2},...,f_{m}\}\subseteq\{f'_{1},f'_{2},...,f'_{n}\}=F'$.
In Fig.~\ref{fig:A-concept-lattice_minsup=00003D00003D00003D00003D00003D00003D00003D00003D00003D2},
$\{u_{A},u_{B}\}\leq\{u_{A},u_{B},u_{C}\}$. The following is easy
to verify:

\begin{lemma} $F\leq F'\rightarrow o(F)\supseteq o(F').$ \end{lemma}

Finally, we bring in the notion of a \textit{concept lattice}, the
lattice defined over the redescription sets using the operator $\leq$
with a join of ${\cal F}$ and a meet of $\emptyset$. Returning once
again to our running example, Figure~\ref{fig:A-concept-lattice_minsup=00003D00003D00003D00003D00003D00003D00003D00003D00003D2}
shows a concept lattice with the lower nodes being relaxations of
the higher nodes in the lattice. In this space we say that a descriptor
$F$ is a child of $F'$ if and only if $F\leq F'$ and $\forall F'':F\leq F''\leq F'\rightarrow(F''=F)\vee(F''=F')$.
Thus in our running example ${\bf C}_{10}$ has two children: ${\bf C}_{4}$
and ${\bf C}_{7}$.

\begin{definition} The \textit{Jaccard coefficient} between two objects
is defined as\vspace{-0.12in}
 \[
J(o_{1},o_{2})=\frac{\left|f(o_{1})\cap f(o_{2})\right|}{\left|f(o_{1})\cup f(o_{2})\right|}\]

\vspace{-0.12in}
\end{definition}

The Jaccard coefficient is thus a measure of how similar two objects
are based upon how many features they share in proportion to their
overall size.

\begin{definition} The \textit{Soergel distance} between two objects
$o_{1}$ and $o_{2}$ is given by\vspace{-0.1in}
\[
D(o_{1},o_{2})=\frac{|f(o_{1})|+|f(o_{2})|-2|f(o_{1})\cap f(o_{2})|}{|f(o_{1})|+|f(o_{2})|-|f(o_{1})\cap f(o_{2})|}\]
\vspace{-0.25in}
 \end{definition}

Unlike the Jaccard coefficient (which is a similarity measure), the
Soergel distance is a true distance measure: it is exactly 0.0 when
the objects $o_{1}$ and $o_{2}$ have exactly the same features,
is symmetric, and obeys the triangle inequality. We also note that
the Soergel distance is exactly 1.0 when the induced feature sets
are disjoint. 

We now define two types of paths---clique paths and hammock paths---which
we use to constrain our path generation.

\begin{definition} \label{hammock-def} A \textit{hammock} of width
$w$ is a tuple $(o_{1},o_{2})\in{\cal O}\times{\cal O}$ where $|f(o_{1})\cap f(o_{2})|=w$.
\end{definition}

A hammock is thus a pair of objects which share at least $w$ common
features. Another way to think of a hammock is as a redescription
set containing two objects ($\{o_{1}\}$ and $\{o_{2}\}$), with a
feature set which contains at least $w$ features. Observe that a
hammock is a basic unit of similarity modeling in recommender systems
and many other applications: it posits similarity in one domain using
relationships with another domain. 
We now define hammock paths composed of hammocks.

\begin{definition} \label{hammockpath-def} A \textit{hammock path}
with width $w$, from object $o_{1}$ to object $o_{t}$, denoted
by $H^{w}(o_{1},o_{t})$, is a series of objects $\langle o_{1},o_{2},...,o_{t-1},o_{t}\rangle$
such that $\forall i:1\leq i\leq t-1$ the pair $(o_{i},o_{i+1})$
is a hammock of width at least $w$. \end{definition}

We now define a \textit{clique} which extends the concept of a hammock
to a set of objects rather than a single pair:

\begin{definition} \label{clique-def} A $k$-clique, $q^{k,w}$,
is a set of $k$ objects $O=\{o_{1},o_{2},\ldots,o_{k}\}$ such that
$\forall o_{i},o_{j}\in O:\,|f(o_{i})\cap f(o_{j})|\geq w$, for some
width $w$. The function $\Upsilon$ maps a clique to the set of $k$
objects contained in said clique. \end{definition}

Note that a $k$-clique is defined over a collection of $k(k-1)/2$
hammocks of width at least $w$; the overlap threshold applied to
every pair of objects of a clique ensures that there is at least as
high an overlap between any two objects in the clique.

\begin{definition} \label{cliquepath-def} A $k$-clique path $Q^{k,w}(o_{1},o_{t})$
is a series of $t-1$ consecutive $k$-cliques $Q=\langle q_{1},q_{2},\ldots,q_{t-1}\rangle$
such that $o_{1}\in\Upsilon(q_{1})$, $o_{t}\in\Upsilon(q_{t-1})$,
and $\forall q_{i}\in Q:\Upsilon\left(q_{i}\right)\cap\Upsilon\left(q_{i+1}\right)\neq\emptyset$.
\end{definition}

The problem we seek to address can now be given as a pair of constraints
upon the clique and hammock paths between two objects, formally defined
as follows:

\begin{problem} Given ${\cal D}=({\cal O},{\cal F})$, start and
end objects $o_{1}\in{\cal O}$ and $o_{t}\in{\cal O}$, we seek to
find a chain/series of objects $S_{1,t}=\langle o_{1},o_{2},\ldots,o_{t}\rangle$
where $S_{1,t}$ is a hammock path $H^{w}(o_{1},o_{t})$ for some
width $w$, and further, there is some $k$-clique path $Q^{k,w}(o_{1},o_{t})=\langle q_{1},q_{2},\ldots,q_{t-1}\rangle$
where $\forall i:1\leq i\leq t-1$, $o_{i}\in\Upsilon(q_{i})$. \end{problem}

\vspace{-0.1in}

\section{Extension to Vector Spaces}

\vspace{-0.15in}

To accommodate vector spaces, we generalize Definition~\ref{Def_database}
so that a database ${\cal D}=({\cal O},{\cal F},V)$ is now a tuple
defined over a set of objects ${\cal O}$, features ${\cal F}$, and
a function $V:{\cal O}\times{\cal F}\rightarrow\mathbb{R}$. This
helps us go beyond simple binary associations to record the strength
of the association (e.g., rating values, term weights) or other auxiliary
continuous-valued data. (Our running example of Figure~\ref{fig:A-concept-lattice_minsup=00003D00003D00003D00003D00003D00003D00003D00003D00003D2}
contains movie ratings by users.) The \textit{weighted Soergel distance}
between two objects $o_{1}$ and $o_{2}$ is defined by\vspace{-0.1in}
\[
\mathbb{D}(o_{1},o_{2})=\frac{{\displaystyle \sum_{f\in{\cal F}}|V(o_{1},f)-V(o_{2},f)|}}{{\displaystyle \sum_{f\in{\cal F}}\max(V(o_{1},f),V(o_{2},f))}}\]
\vspace{-0.15in}

which reduces to the unweighted case if $V(o,f)=1$ for any object
$o\in o(f)$ and $V(o,f)=0$ for any object $o\not\in o(f)$. Definitions~\ref{hammock-def},~\ref{hammockpath-def},~\ref{clique-def},~\ref{cliquepath-def}
get similarly generalized. In place of the hammock width constraint
$w$, we use a weighted Soergel distance threshold $\theta$ that
must be satisfied between the vectors that aim to form hammocks, hammock
paths, cliques, or clique paths. In place of the notation $H^{w}(o_{1},o_{t})$,
we use $H^{\theta}(o_{1},o_{t})$, and so on. 
The new problem becomes:

\begin{problem} Given ${\cal D}=({\cal O},{\cal F},V)$, start and
end objects $o_{1}\in{\cal O}$ and $o_{t}\in{\cal O}$, we seek to
find a chain/series of objects $S_{1,t}=\langle o_{1},o_{2},\ldots,o_{t}\rangle$
where $S_{1,t}$ is a hammock path $H^{\theta}(o_{1},o_{t})$ for
some distance $\theta$, and further, there is some $k$-clique path
$Q^{k,\theta}(o_{1},o_{t})=\langle q_{1},q_{2},\ldots,q_{t-1}\rangle$
where $\forall i:1\leq i\leq t-1$, $o_{i}\in\Upsilon(q_{i})$. \end{problem}

\vspace{-0.1in}
 \Section{Algorithms} \vspace{-0.15in}

Our overall methodology is based on using the concept lattice to structure
the search for paths. Recall that the two parameters influencing the
quality of the path---hammock width and clique size---impose a duality.
The hammock width is posed over feature sets but the clique size is
over objects. We use the clique size to prune the concept lattice
during construction (by incorporating it as a support constraint)
and the hammock width to select candidates for dynamic construction
of paths. There are three key computational stages: (i) construction
of the concept lattice, (ii) generating promising candidates for path
following, and (iii) evaluating candidates for potential to lead to
destination. 
Of these, the first stage can be viewed as a startup cost that can
be amortized over multiple path finding tasks.

We adopt the CHARM-L~\cite{Zaki_Reasoning} algorithm of Zaki for
constructing concept lattices and mining redescriptions. The second
and third stages are organized as part of an A{*} search algorithm
that begins with the starting object, uses the concept lattice to
identify candidates satisfying the hammock and clique size requirements,
and evaluates them heuristically for their promise in leading to the
end object. In practice, we will place a limit on the branching factor
($b$) of the search, thus sacrificing completeness for efficiency.
We showcase these steps in detail below, including the construction
of admissible heuristics.

\vspace{-0.1in}
 \SubSection{Successor Generation} \vspace{-0.1in}

Successor generation is the task of, given an object, using the hammock
and clique size requirements, to identify a set of possible successors
for path following. Note that this does not use the end object in
its computation. We study three techniques for successor generation:

\itemsep=1pt \parsep=5pt 
\begin{enumerate}
\item Cover Tree Nearest Neighbor, 
\item Nearest Neighbors Approximation (NNA), and 
\item $k$-Clique Near Neighbor ($k$CNN). 
\end{enumerate}
The first technique is targeted toward finding paths when the clique
size requirement is 2 (top and middle paths of Fig.~\ref{fig:Movie-recommendationsExample}).
That is, this technique is able to generate $b$ singleton successors,
where $b$ is the branching factor of the search. The second two techniques
concentrate on paths of any clique size, such as the bottom path of
Fig.~\ref{fig:Movie-recommendationsExample}. Instead of generating
singleton successors, the NNA and $k$CNN algorithms are able to generate
successor-sets, where each of these sets constitutes a candidate $k$-clique
with the given object.


\subsubsection{Cover Tree Nearest Neighbor}

The cover tree~\cite{Beygelzimer_Cover} is a data structure for
fast nearest neighbor operations in a space of objects organized alongside
any distance metric (here, we use the Soergel distance \cite{Leach_An}
\comment{ \narenc{cite the Soergel distance equation or equations.}}).
The space complexity is $O(\|{\cal O}\|)$, i.e., linear in the object
size of the database. A nearest neighbor query requires logarithmic
time in the object space $O\left(c^{12}\text{log}\left(n\right)\right)$
where $c$ is the expansion constant associated with the featureset
dimension of the dataset (see~\cite{Beygelzimer_Cover} for details).


\subsubsection{Nearest Neighbors Approximation (NNA)}

The second mechanism we use for successor generation is to approximate
the nearest neighbors of an object using the concept lattice. 
We use the Jaccard coefficient between two objects as an indicator
to inversely (and approximately) track the Soergel distance between
the objects. In order to efficiently calculate an object's nearest
neighbors, however, we cannot simply calculate the Jaccard coefficient
between it and every other object. We harness the concept lattice
to avoid wasteful comparisons. We first define a predicate $\tau$
on redescriptions and objects.

\begin{algorithm}
\begin{algorithmic} \REQUIRE An object $o\in{\cal O}$

\STATE ${\it fringe}\gets\top(o)$ order by feature set size

\STATE ${\it prospects}\gets\emptyset$

\WHILE{${\it fringe}\neq\emptyset$}

\STATE $r\gets$ dequeue from \textit{fringe} \WHILE{${\it prospects}\neq\emptyset$}

\STATE $o'\gets$ head \textit{prospects}

\IF{$J(o,o')>\frac{|{\rm Items}(o)|}{|{\rm Items}(r)|}$}

\STATE yield $o'$

\STATE dequeue \textit{prospects}

\ELSE

\STATE break

\ENDIF

\ENDWHILE \FORALL{$r'\in$ ChildrenOf $r$} \STATE add $r'$
to \textit{fringe} order by feature set size \FORALL{$o'$ in $o(r')$}
\STATE add $o'$ to \textit{prospects} order by $J(o,o')$ \ENDFOR
\ENDFOR \ENDWHILE \end{algorithmic}\vspace{-0.05in}

\caption{NNA($o$)}

\label{alg:NNA_algo} 
\end{algorithm}

\begin{definition} For redescription $R_{F}=(O,F)$ and object $o$,
$\tau(o,R_{F})$ if and only if $o\in o(F)$ and there is no $F'$
such that $F\leq F'$ and $o\in o(F')$. \end{definition}

Informally, we can say $\tau(o,R_{F})$ if $F$ is on the edge of
redescriptions which contain the object $o$. In Figure~\ref{fig:A-concept-lattice_minsup=00003D00003D00003D00003D00003D00003D00003D00003D00003D2},
using object $m_{1}$ the only feature set $F$ for which $\tau(m_{1},R_{F})$
would evaluate to true is $F=\{u_{A},u_{D},u_{E}\}$. Note that if
no support threshold (clique constraint) is given, there will always
be exactly one such feature set for each object. When using a support
threshold greater than one, this is no longer true; so we now define
the set $\top(o,2^{{\cal F}})$ for object $o$.

\begin{definition} For object $o$ and all redescriptions $2^{{\cal F}}$
in a concept lattice containing $o$, $\top(o,2^{{\cal F}})=\{F:F\in2^{{\cal F}}\wedge\tau(o,R_{F})\}$.
\end{definition}

We generally omit the $2^{{\cal F}}$ where it is clear from context.
The set $\top(o)$ is then the set of all redescriptions which form
the upper edge in the concept lattice where $o$ appears. All the
objects in Figure~\ref{fig:A-concept-lattice_minsup=00003D00003D00003D00003D00003D00003D00003D00003D00003D2}
have singleton sets for $\top$, however, if we change the support
threshold from one object to five objects, then the object $m_{1}$
will have $\top(m_{1})=\{\{u_{A}\},\{u_{D},u_{E}\}\}$.

A formal description of our NNA algorithm is shown in Algorithm~\ref{alg:NNA_algo}.


\begin{definition} ${\rm NNA}_{k}(o)$ is the $k^{{\rm th}}$ object
returned by the NNA algorithm when run on input object $o$. \end{definition}

\begin{theorem} Given a database with object set ${\cal O}$ with
size $|{\cal O}|=n$ containing object $o\in{\cal O}$, $\forall i,j\in{\mathbb{Z}}:1\leq i\leq j\leq n\rightarrow J(o,{\rm NNA}_{i}(o))\geq J(o,{\rm NNA}_{j}(o))$.
\end{theorem}

(Proof omitted due to space constraints.) In other words, NNA returns
better approximate redescriptions of an object first. This is still,
however, an approximation since it uses the Jacccard coefficient rather
than the Soergel distance.


\subsubsection{$k$-Clique Near Neighbor ($k$CNN)}


\comment{The algorithms presented thus far have been oriented toward
finding paths with a clique size of 2 (e.g., top two paths of Fig.~\ref{fig:Movie-recommendationsExample}).
We now turn our attention to enforcing clique size requirements.}
The basic idea of the $k$CNN approach is, in addition to finding
a good set of successor nodes for a given object $o$, to be able
to have sufficient number of them so that, combinatorially, they contribute
a desired number of cliques. With a clique size constraint of $k$,
it is not sufficient to merely pick the top $k$ neighbors of the
given object, since the successor generation function expects multiple
clique candidates. (Note that, even if we picked the top $k$ neighbors,
we will still need to subject them to a check to verify that every
pair satisfies the hammock width constraint.)\comment{ \narenc{Don't
keep saying `hammock size'. It is `hammock width'. Be consistent.
Do a string replace uniformly.}} Given that this function expects
$b$ clique candidates, 
minimum 
number $m$ of candidate objects to identify can be cast as the solution
to the inequalities:

 \[
\left(\begin{array}{c}
m-1\\
k\end{array}\right)<b\,\text{\, and}\,\,\left(\begin{array}{c}
m\\
k\end{array}\right)\geq b\]

The object list of each concept of the lattice is ordered in the number
of features (e.g., see Fig.~\ref{fig:A-concept-lattice_minsup=00003D00003D00003D00003D00003D00003D00003D00003D00003D2})
and this aids in picking the top $m$ candidate objects for the given
object $o$. We pick up these $m$ candidate objects for $o$ from
the object list of the concept containing the longest feature set
and redescription set containing $o$. Note that, in practice, the
object list of each concept is much larger than $m$ and as a result
$k$CNN does not need to traverse the lattice to obtain promising candidates.
$k$CNN thus forms combinations of size $k$
from these $m$ objects to obtain a total of $b$ $k$-cliques. Since
$m$ is calculated using the two inequalities, the total number of
such combinations is equal to or slightly greater than $b$ (but never
less than $b$). Each clique is given an average distance score calculated
from the distances of the objects of the clique and the current object
$o$. This aids $k$CNN in returning a priority queue of exactly $b$
candidate $k$-cliques.

 \SubSection{Evaluating Candidates} 

We now have a basket of candidates that are close to the current object
and must determine which of these has potential to lead to the destination.
We present two operational modes to rank candidates: (i) the normal
mode and (ii) the mixed mode. 


 %
\begin{figure}
\centering\includegraphics[scale=0.3]{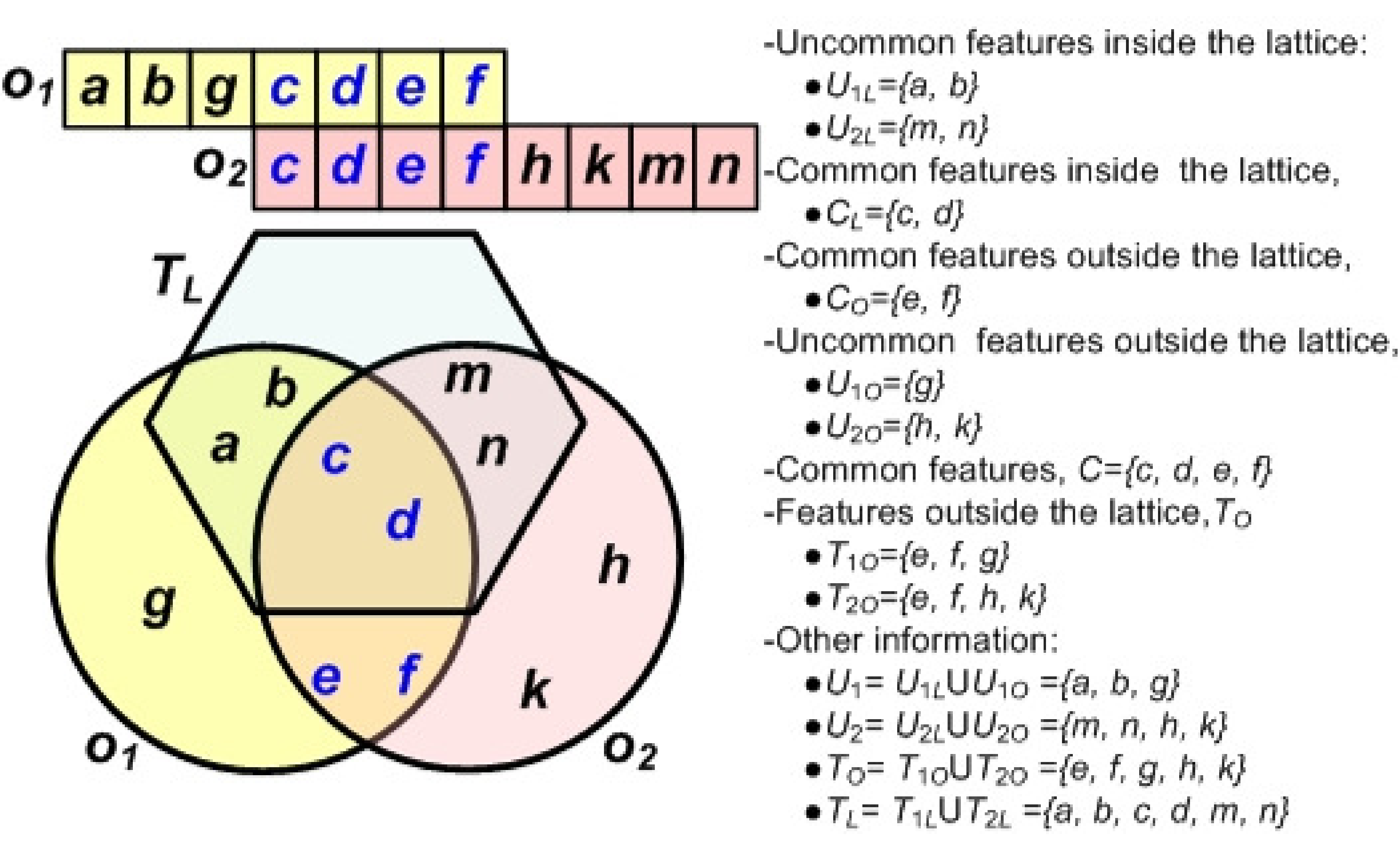}

\caption{\small{Distribution of common and uncommon features of objects $o_{1}$ and
$o_{2}$ inside and outside the concept lattice. (The hexagon indicates
the concept lattice).}}


\label{fig:Distribution-of-common_uncommon}


\end{figure}

\subsubsection{Normal Mode}

The normal mode is suitable for the general case where we have all
the objects and features resident in the database. The primary criteria
of optimality for the A{*} search procedure is the cumulative Soergel
distance of the path. We use the straight line Soergel distance for
the heuristic. It is easy to see that this can never oveestimate the
cost of a path from 
any object $o$ to the goal (and is hence admissible), because the
Soergel distance maintains the triangle inequality. 

 %
\begin{figure*}[b]
\centering

$\mathbb{D}_{\text{mixed}}(o_{1},o_{2})=\frac{\left|U_{1L}\right|\times\text{minw}(o_{1})+\left|U_{2L}\right|\times\text{minw}(o_{2})+\sum_{f\in C_{L}}|V(o_{1},f)-V(o_{2},f)|}{\left|U_{1L}\right|\times\text{minw}(o_{1})+\left|U_{2L}\right|\times\text{minw}(o_{2})+\sum_{f\in C_{L}}\max(V(o_{1},f),V(o_{2},f))+|T_{O}|\times\max(\text{maxw}(o_{1}),\text{maxw}(o_{2}))}$

\caption{\small{The mixed mode Soergel distance equation.} \comment{\narenc{Why
use f' as the index variable when you can use f? Keep things simple}}}

\label{fig:Mixed-Soergel_Formula}


\end{figure*}

\subsubsection{Mixed Mode\label{sub:Mixed-Mode-Approach}}

The mixed mode distance measure is effective 
for large datasets where only important information are stored but
other information are removed from the system after recording some
of their aggregated information, to save space and cost of establishment.
With the mixed mode approach, for simplicity, we assume that all the
information about items outside the concept lattice are absent but
some of their aggregated information like number of features truncated
are provided. Fig.~\ref{fig:Distribution-of-common_uncommon} shows
the distribution of common and uncommon features of objects $o_{1}$
and $o_{2}$ inside and outside a concept lattice. The mixed mode
Soergel distance is given by:

\comment{\narenc{put the underscore of `mixed' right after the
D, before the arguments. Else it looks funny. It looks as if `mixed'
is a subscript for the tuple (o1,o2).}} 
 \[
D_{\text{mixed}}(o_{1},o_{2})=\frac{\left|U_{1L}\right|+\left|U_{2L}\right|}{\left|U_{1L}\right|+\left|U_{2L}\right|+\left|C_{L}\right|+\left|T_{O}\right|}\]
 where the terms are defined in Fig.~\ref{fig:Distribution-of-common_uncommon}.

\comment{{\color{red} Michael: $U$ is never defined} }

\begin{theorem} $D_{\text{mixed}}(o_{1},o_{2})$ never overestimates
the original Soergel distance $D(o_{1},o_{2})$. \end{theorem}

\emph{Proof:} Omitted due to more general Theorem~\ref{mmtheorem}
later.


\subsubsection{Mixed Mode in Vector Spaces}

However, the mixed mode becomes more complex in the vector space model
than its set model. Figure \ref{fig:Mixed-Soergel_Formula} shows
our formula for the mixed mode in the vector space. Similar to the
unweighted one, we assume that all the information about features
of objects outside the concept lattice are absent in the weighted
mixed mode approach. But some of their aggregated information like
minimum (minw) and maximum (maxw) weights are provided. 

Consider the set of features $T_{O}$ that do not appear in the lattice
due to the support threshold of the concept lattice, \emph{minsup}.
Some features of $T_{O}$ can be common to both objects $o_{1}$ and
$o_{2}$. $\left|U_{1O}\right|$ and $\left|U_{2O}\right|$ are the
numbers of uncommon features in objects $o_{1}$ and $o_{2}$, which
are thus outside the frequent concept lattice. Length $\left|T_{O}\right|$
is a known variable due to the recorded aggregated information, but
$\left|U_{1O}\right|$ and $\left|U_{2O}\right|$ are unknown. This
is why $\left|U_{1O}\right|$ and $\left|U_{2O}\right|$ do not appear
in $\mathbb{D}_{\text{mixed}}(o_{1},o_{2})$. For $\mathbb{D}_{\text{mixed}}(o_{1},o_{2})$,
we consider that all the features of $T_{O}$ (i.e., features outside
the lattice) are common in both objects $o_{1}$ and $o_{2}$ and
all these features have the same weight which is $\max(\text{maxw}(o_{1}),\text{maxw}(o_{2}))$.

\begin{theorem} \label{mmtheorem} $\mathbb{D}_{\text{mixed}}(o_{1},o_{2})$
never overestimates the original Soergel distance $\mathbb{D}(o_{1},o_{2})$.
\end{theorem}

\emph{Proof:} Let the numerator of $\mathbb{D}(o_{1},o_{2})$ and
$\mathbb{D}_{\text{mixed}}(o_{1},o_{2})$ be $\eta^{1,2}$ and $\eta_{\text{mixed}}^{1,2}$
respectively. Similarly, let $\varpi^{1,2}$ and $\varpi_{\text{mixed}}^{1,2}$
be the denominators of $\mathbb{D}(o_{1},o_{2})$ and $\mathbb{D}_{\text{mixed}}(o_{1},o_{2})$.
It is clear that $\eta_{\text{mixed}}^{1,2}\leq\eta^{1,2}$. Therefore,
it suffices to show that $\varpi_{\text{mixed}}^{1,2}\geq\varpi^{1,2}$.

For ease of derivation, we define the following notation \comment{
\narenc{Put these in an eqnarray{*} environment so that athe equal
signs are aligned.} \narenc{Sometimes you put \LaTeX{} `max', sometimes
you put textual `max'. Be consistent.}}:

\comment{

$\alpha=\left|U_{1L}\right|\times\text{minw}\left(o_{1}\right)+\left|U_{2L}\right|\times\text{minw}\left(o_{2}\right)$

$\beta=\sum_{f\in C_{L}}\max(V(o_{1},f),V(o_{2},f))$

$\zeta=\sum_{f\in\left(T_{O}-U_{1O}-U_{2O}\right)}\max(V(o_{1},f),V(o_{2},f))$

$\chi=\left|U_{1}\right|\times\text{minw}\left(o_{1}\right)+\left|U_{2}\right|\times\text{minw}\left(o_{2}\right)$

$\xi=\left|U_{1O}\right|\times\text{minw}\left(o_{1}\right)+\left|U_{2O}\right|\times\text{minw}\left(o_{2}\right)$

$\varrho=\left|T_{O}\right|\times\max\left(\text{maxw}\left(o_{1}\right),\text{maxw}\left(o_{2}\right)\right)$

}


\begin{eqnarray*}
\alpha & = & \left|U_{1L}\right|\times\text{minw}\left(o_{1}\right)+\left|U_{2L}\right|\times\text{minw}\left(o_{2}\right)\\
\beta & = & \sum_{f\in C_{L}}\max(V(o_{1},f),V(o_{2},f))\\
\zeta & = & \sum_{f\in\left(T_{O}-U_{1O}-U_{2O}\right)}\max(V(o_{1},f),V(o_{2},f))\\
\chi & = & \left|U_{1}\right|\times\text{minw}\left(o_{1}\right)+\left|U_{2}\right|\times\text{minw}\left(o_{2}\right)\\
\xi & = & \left|U_{1O}\right|\times\text{minw}\left(o_{1}\right)+\left|U_{2O}\right|\times\text{minw}\left(o_{2}\right)\\
\varrho & = & \left|T_{O}\right|\times\max\left(\text{maxw}\left(o_{1}\right),\text{maxw}\left(o_{2}\right)\right)\end{eqnarray*}

\noindent Therefore we have, $\varpi^{1,2}=\chi+\beta+\zeta$. Now,
$\varpi_{\text{mixed}}^{1,2}=\alpha+\beta+\varrho=\chi+\beta+\zeta-\xi+\varrho-\zeta=\varpi^{1,2}+\varrho-\zeta-\xi=\varpi^{1,2}+L$,
where $L=\varrho-\zeta-\xi$. Since $L\geq0$ is always true, $\varpi_{\text{mixed}}^{1,2}\geq\varpi^{1,2}$.
Therefore, $\mathbb{D}_{\text{mixed}}(o_{1},o_{2})\leq\mathbb{D}(o_{1},o_{2})$. $\Box$
\comment{\narenc{the subscript mixed is again in the wrong place.
Also whenever a proof ends, they place a small box at the end. Lookup
some math books/\LaTeX{} references for it.}}


\vspace{-0.01in}
 \Section{Experimental Results} \vspace{-0.1in}

We present our experimental results on four datasets (see Table~\ref{tab:DatasetDescription})
using a 64-bit Windows Vista Intel Core2 Quad CPU Q9450 @ 2.66GHz
and 8 GB physical memory. 

%
\begin{table}[b]
\caption{Dataset characteristics.}

\label{tab:DatasetDescription} \centering \begin{tabular}{|>{\centering}p{0.5in}|>{\centering}p{0.42in}|>{\centering}p{0.45in}|>{\centering}p{0.62in}|>{\centering}p{0.44in}|}
\hline 
\textbf{\small Dataset}{\small {}}  & \textbf{\small \# obj.}{\small {}}  & \textbf{\small \# feat.}{\small {}}  & \textbf{\small \# relations}{\small {}}  & \textbf{\small Sparsity}{\small {} }\tabularnewline
\hline 
{\small Synthetic}  & {\small 1,000}  & {\small 3,679}  & {\small 430,477}  & {\small 99.88}\tabularnewline
\hline 
{\small Netflix}  & {\small 17,770}  & {\small 480,189}  & {\small 100,480,507}  & {\small 99.98}\tabularnewline
\hline 
{\small PubMed}  & {\small 161,693}  & {\small 133,252}  & {\small 2,226,616}  & {\small 99.99}\tabularnewline
\hline 
{\small Cl. Trials}  & {\small 66,526}  & {\small 75,656}  & {\small 907,759}  & {\small 99.99}\tabularnewline
\hline
\end{tabular}


\end{table}

\begin{figure}
\centering\includegraphics[scale=0.45]{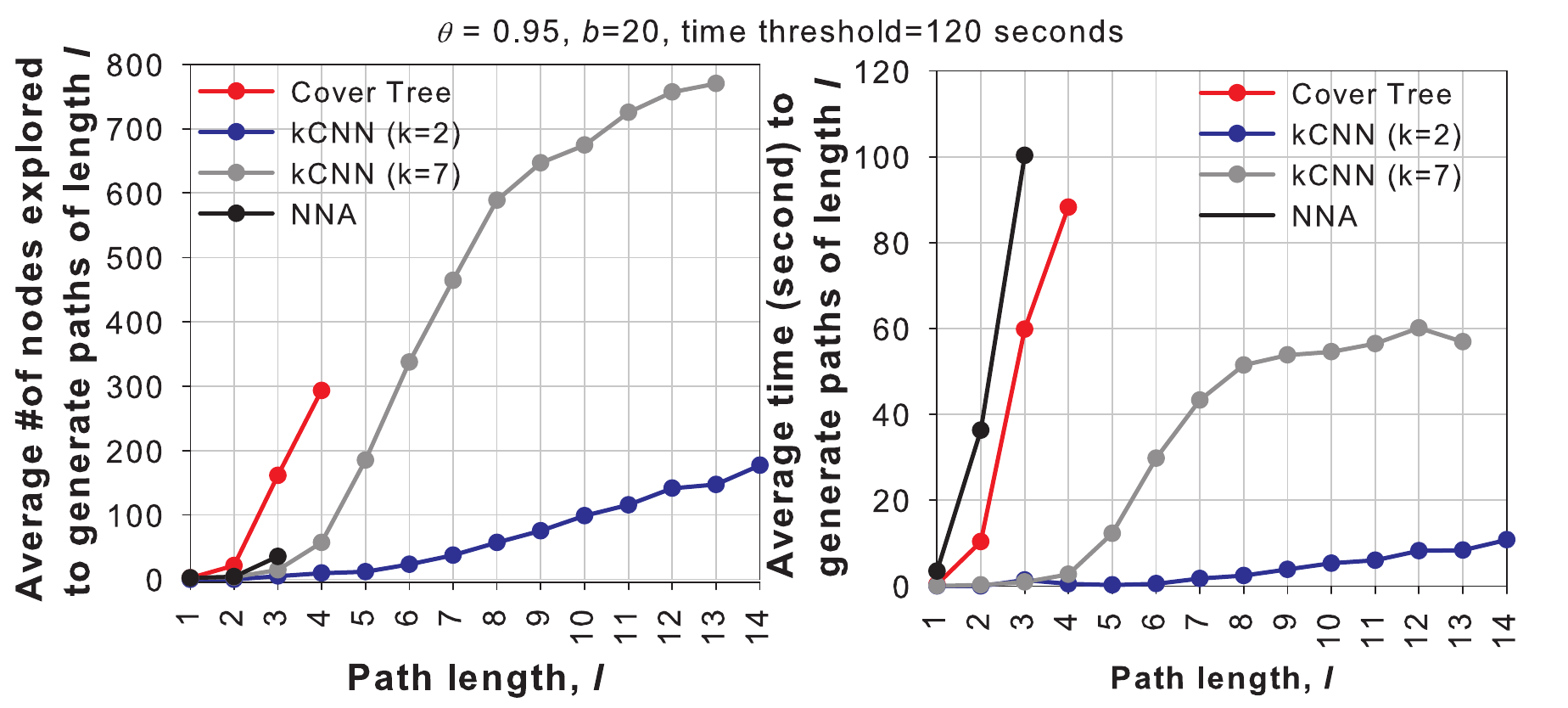}

\caption{\small{Synthetic dataset: Comparison between successor generation strategies.}
\label{fig:Comparison-Netflix1000}}


\end{figure}

\SubSection{Evaluating successor generation strategies} \vspace{-0.1in}
 The goal of our first experiment was to assess the number of nodes
explored by the A{*} search and the time taken as a function of the
discovered path length, as a function of different successor generation
strategies. For this purpose, we defined a synthetic dataset involving
1000 randomly selected movies from the Netflix dataset and aimed to
generate more than 50,000 similarity paths between randomly selected
pairs, with a $\theta$ threshold of 0.95 and a branching factor bound
($b$) of 20. We introduced a time threshold of 120 seconds in the
A{*} search and Fig.~\ref{fig:Comparison-Netflix1000} depicts the
results of only successful searches. 
Around 80\% of these searches failed when using the the cover tree
and NNA approaches, but the $k$CNN approach was able to either successfully
generate paths or to declare that no path exists in less than 120
seconds. As Fig.~\ref{fig:Comparison-Netflix1000} shows, the cover
tree and NNA terminate early due to the applied time constraint but
$k$CNN generated long paths of length 14 and 13 with $k$=2 and 7,
respectively. The runtime trends shown in Fig.~\ref{fig:Comparison-Netflix1000}
(right) also mirror the number of nodes explored in Fig.~\ref{fig:Comparison-Netflix1000}
(left).

This result is not surprising, as the cover tree algorithm does not
factor in the clique constraint, thus preventing it from taking advantage
of the search space reduction that this constraint provides. NNA
does take advantage of this constraint, however, and it
generates a strict ordering on the Jaccard's coefficient
over the cliques, whereas $k$CNN simply generates
some $b$ candidate cliques. In practice, the $k$CNN relaxation
results in the discovery of candidate cliques
more rapidly than the NNA algorithm, while still remaining accurate
as in both cases a post processing step is necessary to determine
if a given candidate does in fact meet the search threshold. Through
the rest of this paper, we thus use the $k$CNN algorithm as it generally
provides the best performance of the three algorithms. 

\vspace{-0.1in}
 \SubSection{Netflix dataset} \vspace{-0.1in}
 Viewing movies as objects and users as features, we construct the
concept lattice for the Netflix dataset with a support constraint
of 20\%. The resulting lattice contains 5,884 concepts, 120 of whom
were leaves and the rest had child concepts. We picked 50,000 pairs
of movies and attempted to generate hammock paths between them.

\begin{figure*}[b]
\centering\includegraphics[scale=0.28]{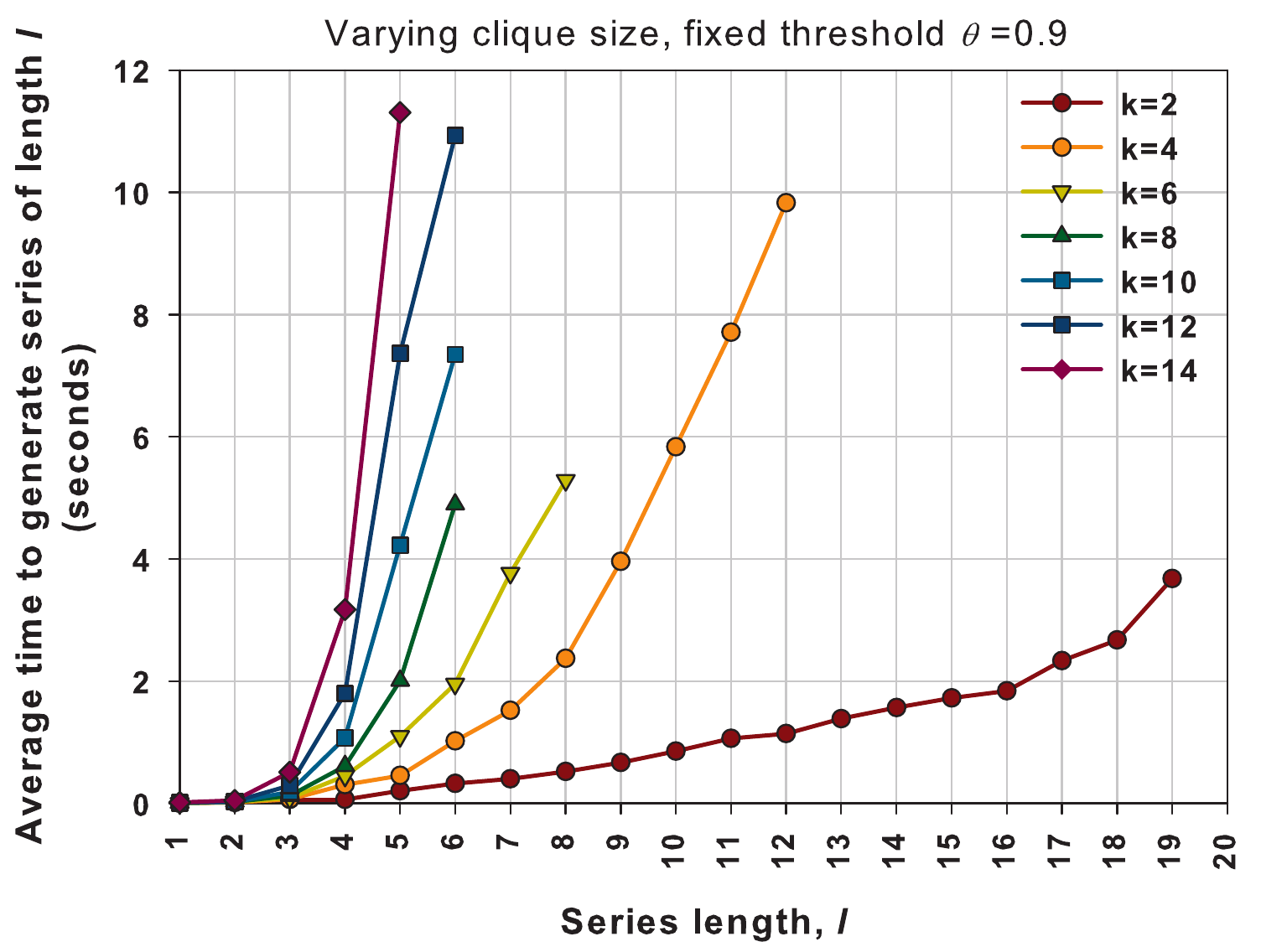}\includegraphics[scale=0.28]{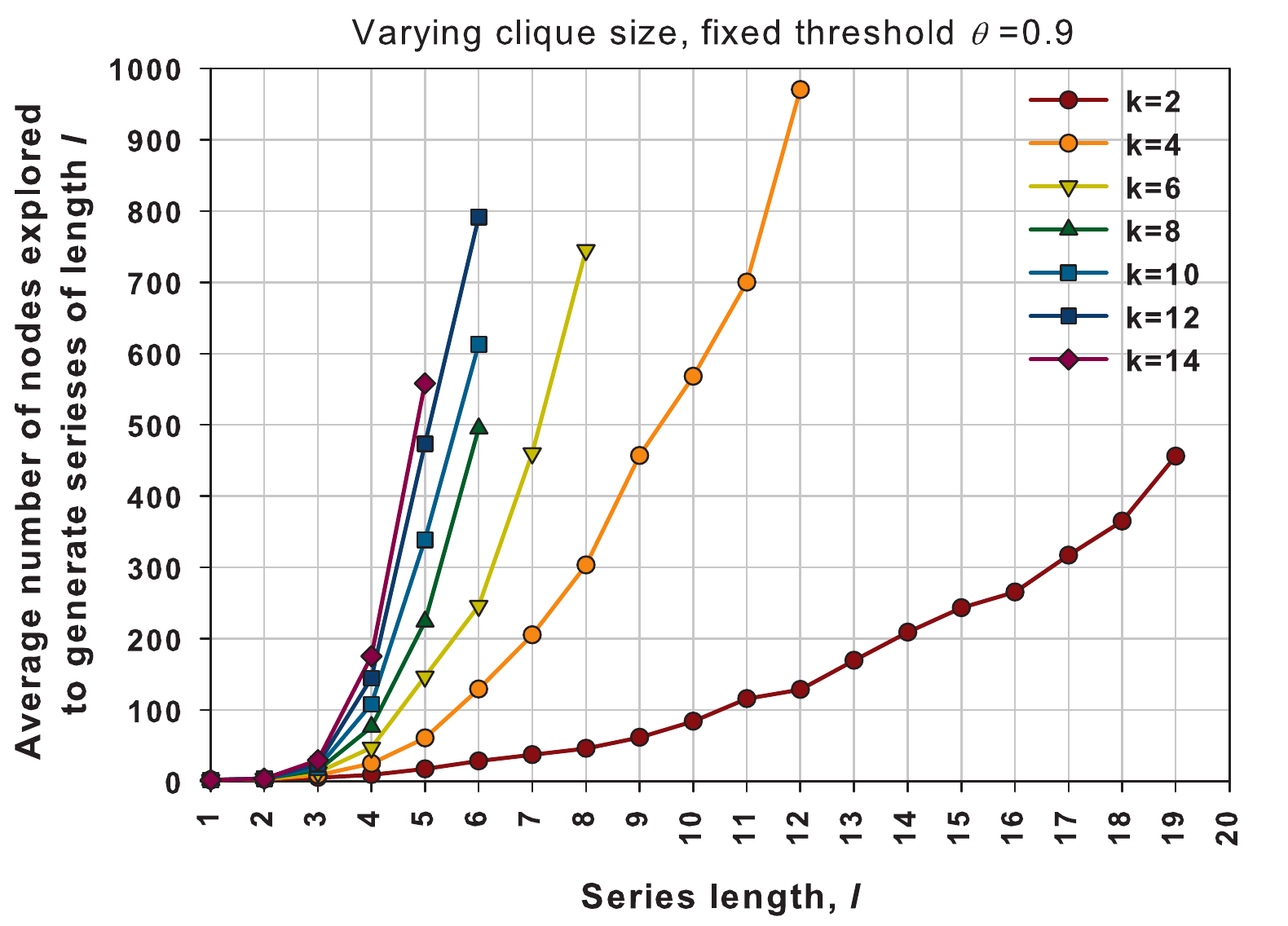}\includegraphics[scale=0.28]{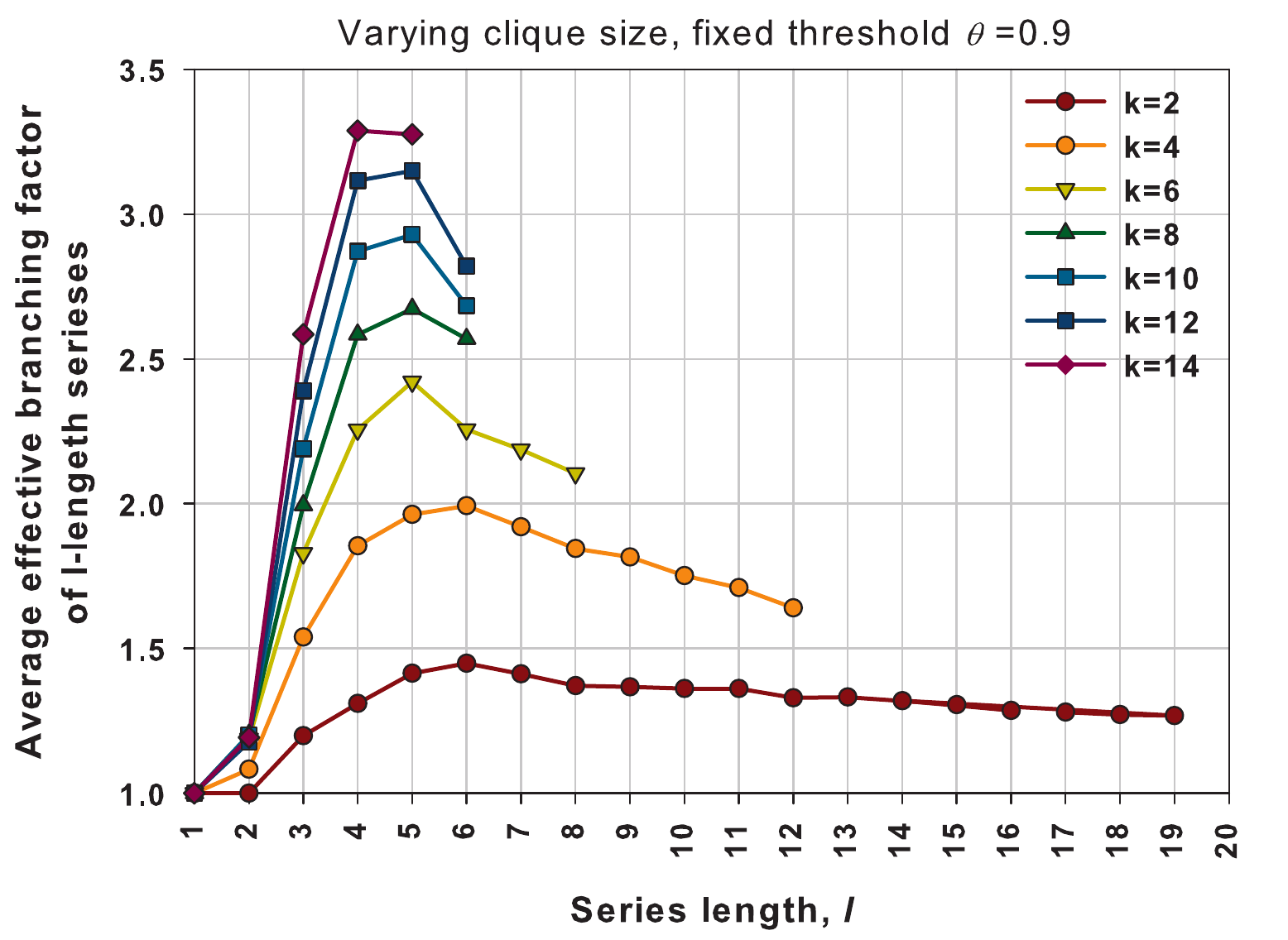}\includegraphics[scale=0.28]{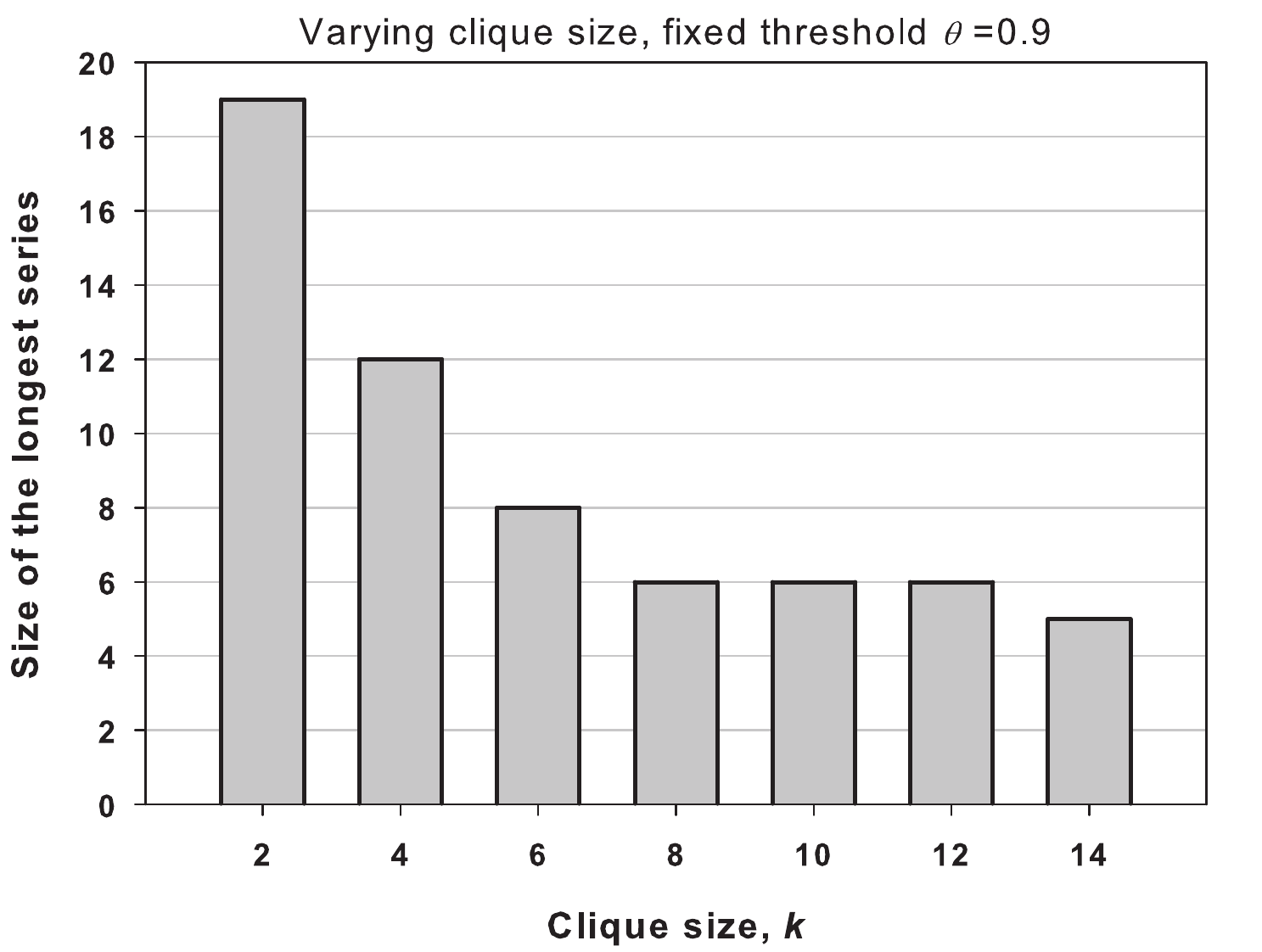}\vspace{-0.1in}

\caption{\small{Netflix: Explorations by varying clique size requirement at a fixed
Soergel threshold ($\theta$=0.9). The first three plots show average
run time, number of nodes explored, and effective branching factor,
as a function of path lengths, for different clique size requirements.
The fourth plot shows that the size of the longest path reduces as
the clique size increases.}\label{fig:Varyingk_fixedthreshold}}


\end{figure*}

\begin{figure*}[!t]
\centering\includegraphics[scale=0.1]{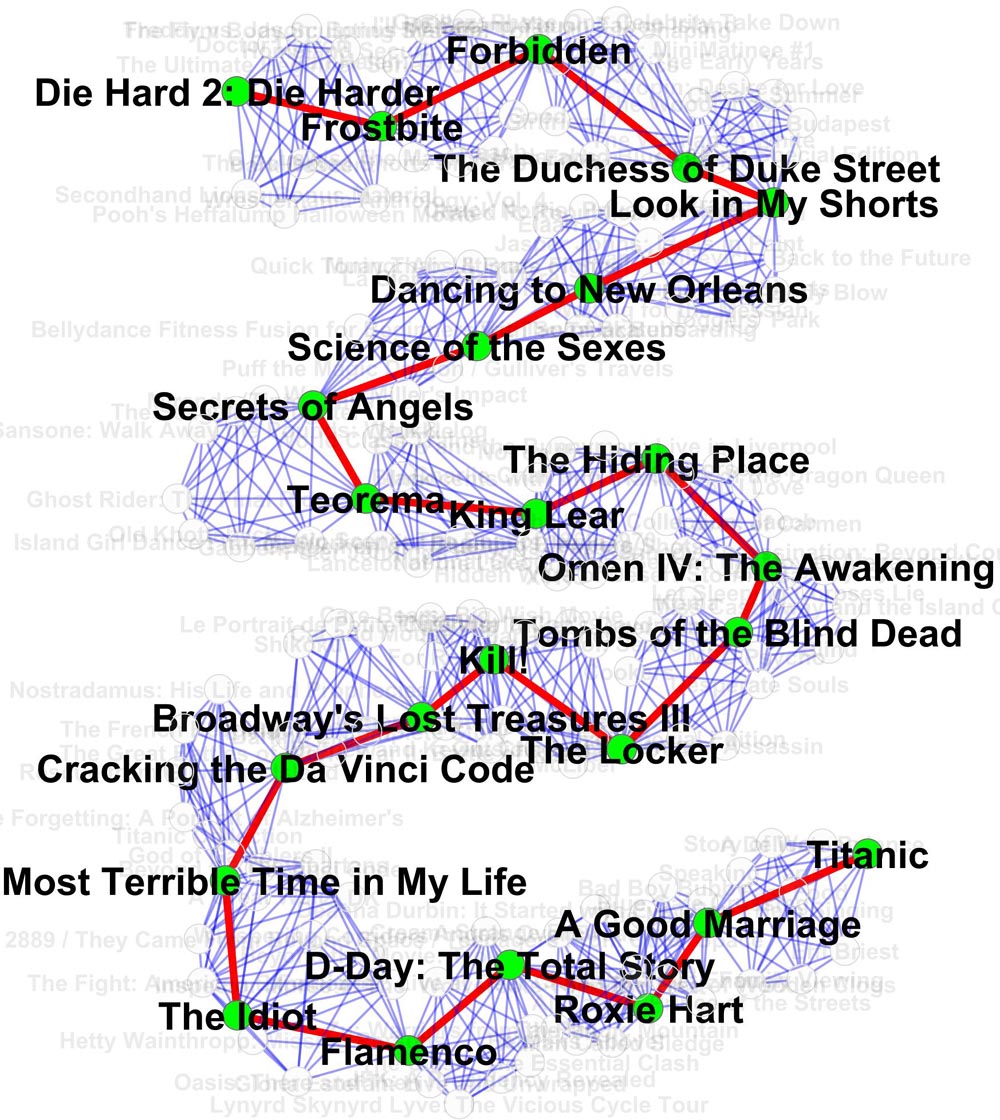}
\includegraphics[scale=0.11]{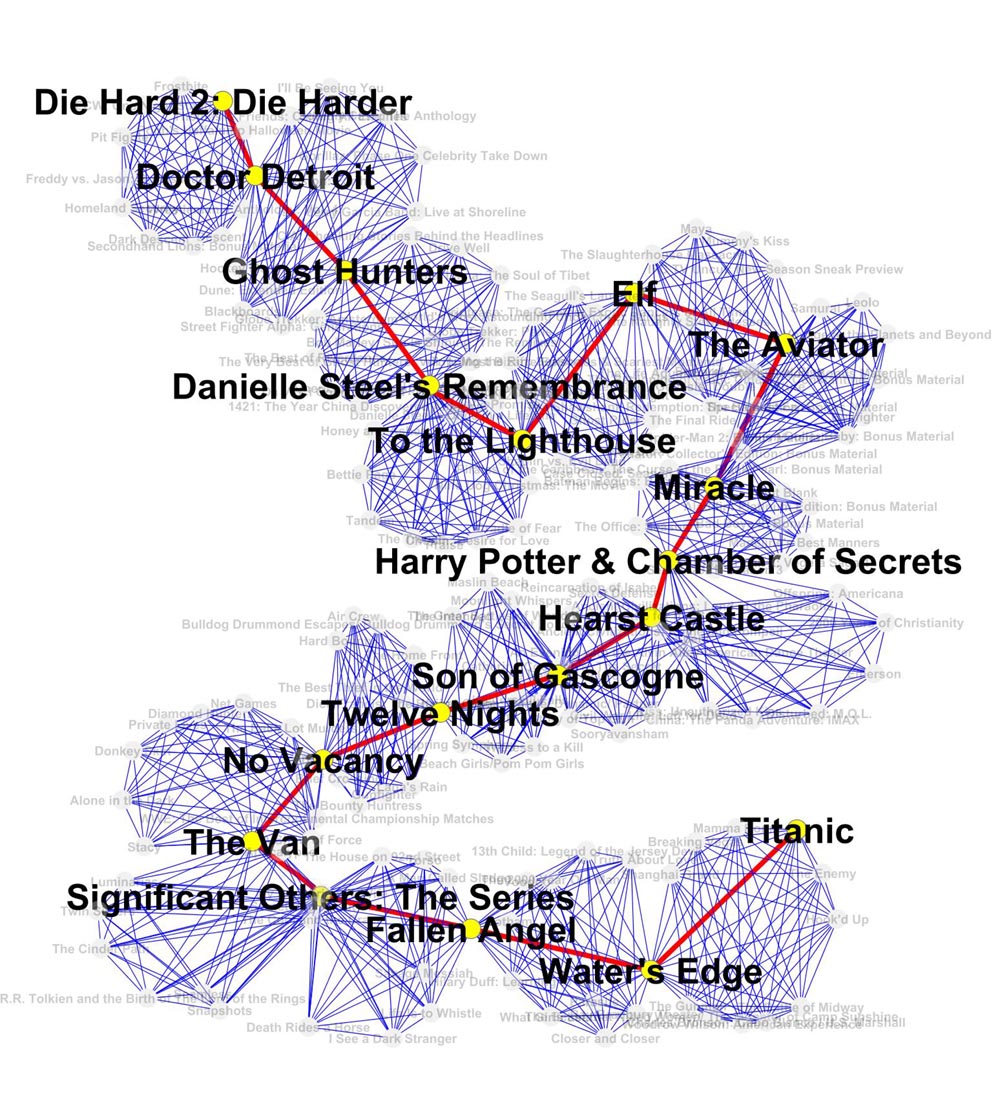}
\includegraphics[scale=0.09]{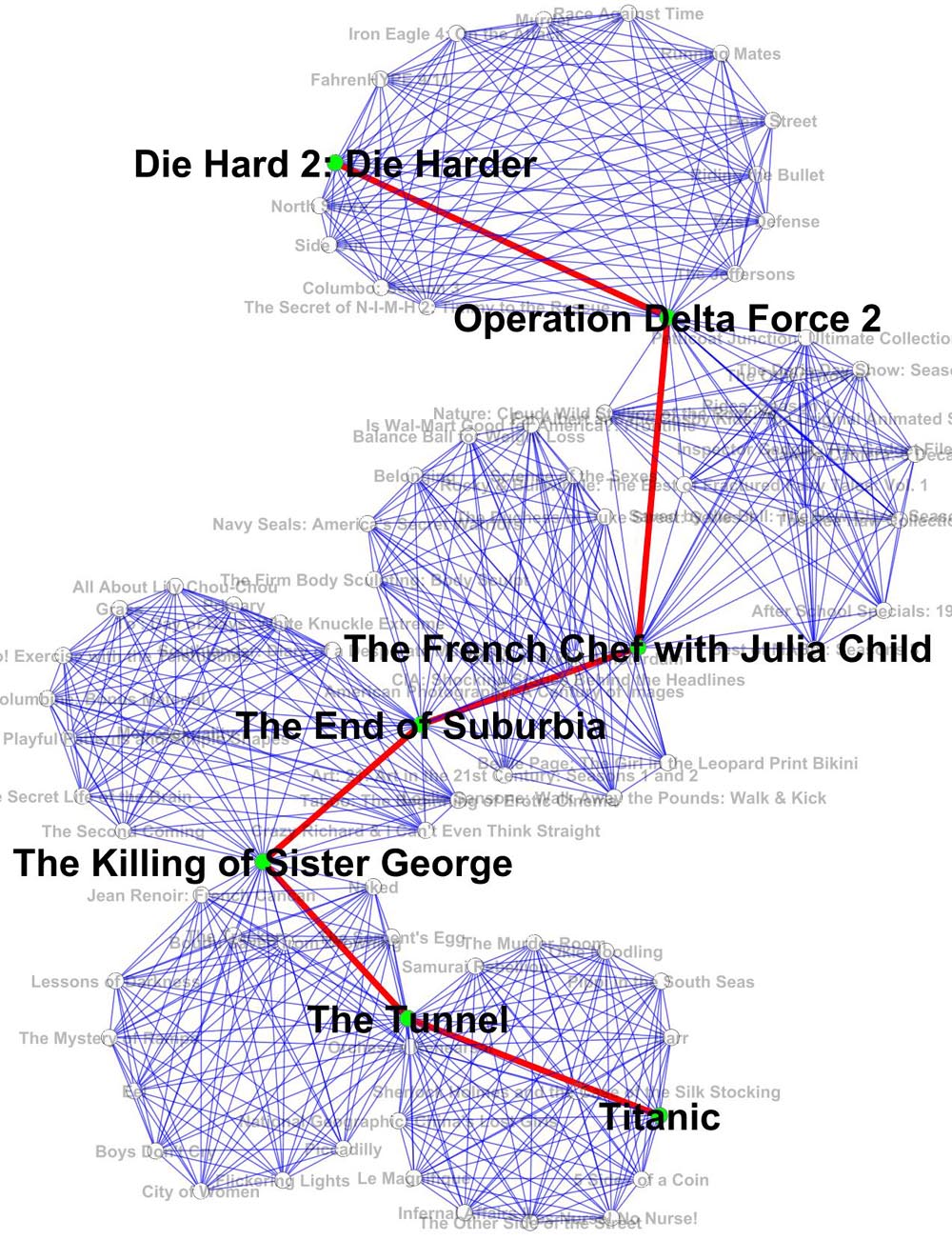}
\includegraphics[scale=0.11]{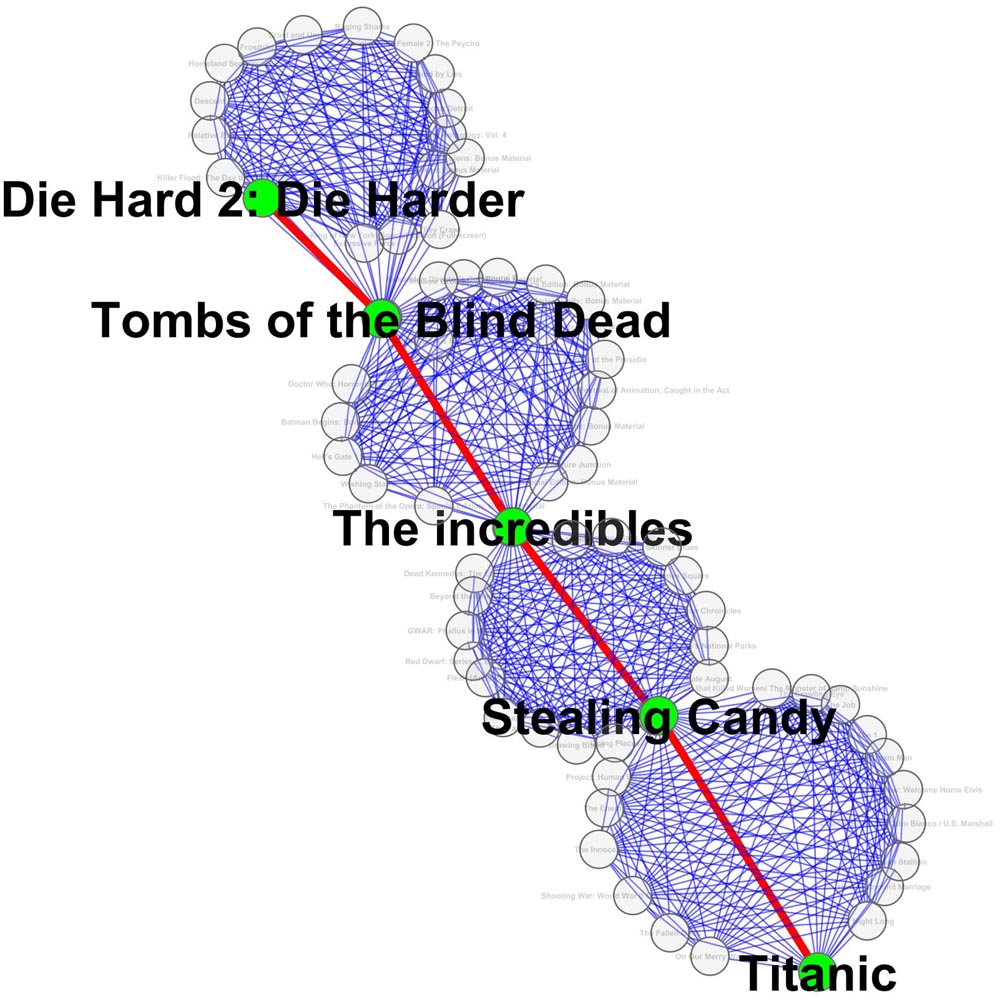}

Clique size=9 ~~~~~~~~~~~~~~~~~~~~~~~Clique
size=12 ~~~~~~~~~~~~~~~~~~~~~~~~Clique size=15~~~~~~~~~~~~~~~~~~~~~~Clique
size=18

\vspace{-0.1in}
\caption{\small{Netflix: Hammock paths from 1943 classic drama {}``Titanic'' to
the action movie {}``Die Hard 2: Die Harder'' for four different
clique size requirements.}\label{fig:TitanicDieHardExample_Example}}


\end{figure*}

Figure \ref{fig:Varyingk_fixedthreshold} shows our experimental results
with varying clique size and fixed distance threshold. From the graph
at the left, it is evident that the performance of our kCNN algorithm
depends on the clique size since the graph shows that hammock paths
with lower clique sizes are generated faster than those with higher
clique sizes. The tendency of the run time basically follows the required
number of nodes explored to generate the paths (the second graph).
The third graph shows the corresponding trends of effective branching
factor. It is evident that the higher the clique size the worse (or
larger) the effective branching factor is. Effective branching factor
is a measure to understand the size of the traversed search space
compared with corresponding breadth first search (BFS). Therefore
it is a measure of the efficacy of the heuristic in pruning out unwanted
paths. Although the effective branching factor becomes worse with
larger clique size, it generates smaller hammock paths. Thus, the
lengths of the paths are also affected by the clique size requirement.
The plot at the right depicts the lengths of the hammock paths as
a function of clique size. It demonstrates that our algorithm has
the capability to generate longer chains with smaller cliques. For
example, Figure~\ref{fig:TitanicDieHardExample_Example} gives hammock
paths between the 1943 classic drama {}``Titanic'' and the action
movie {}``Die Hard 2: Die Harder'', for four different clique sizes
($k$=9, 12, 15 and 18). 

\comment{\textcolor{red}{Discussion: Note that the mixed mode approach
cannot be fairly compared with the distance threshold based approach
because, the mixed mode approach depends on overlap as the threshold
but the other depends on distance threshold. We avoided the actual
distance threshold in the mixed mode approach of kCNN because it is
not possible to convert the Soergel distance threshold provided by
the user to a mixed mode distance threshold from aggregated information
since there can be millions of possible combinations of aggregated
information. Mixed mode distance never over estimates the actual distance.
So, it can be used for successor generation purpose and as a heuristic.
The generated serieses by the mixed mode should be deviated from the
serieses generated by the distance threshold based stories by factors
dependent on how much information is lost during truncating the database.}}

 \SubSection{PubMed Abstracts}

%
\begin{figure*}[!t]
\subfigure[Varying threshold $\theta$, fixed clique size ($k$=6). (Left) Number of nodes explored by the A* procedure. It shows that the
application of lower thresholds results in a lower number of explored nodes, as a function of path length. (Middle) The higher the distance threshold, the higher the run time to generate paths. (Right) Lower distance threshold has the trend to generate paths with low effective branching factor.]{ \label{fig:Varying-threshold-fixedK_PubMed} \centering \includegraphics[scale=0.38]{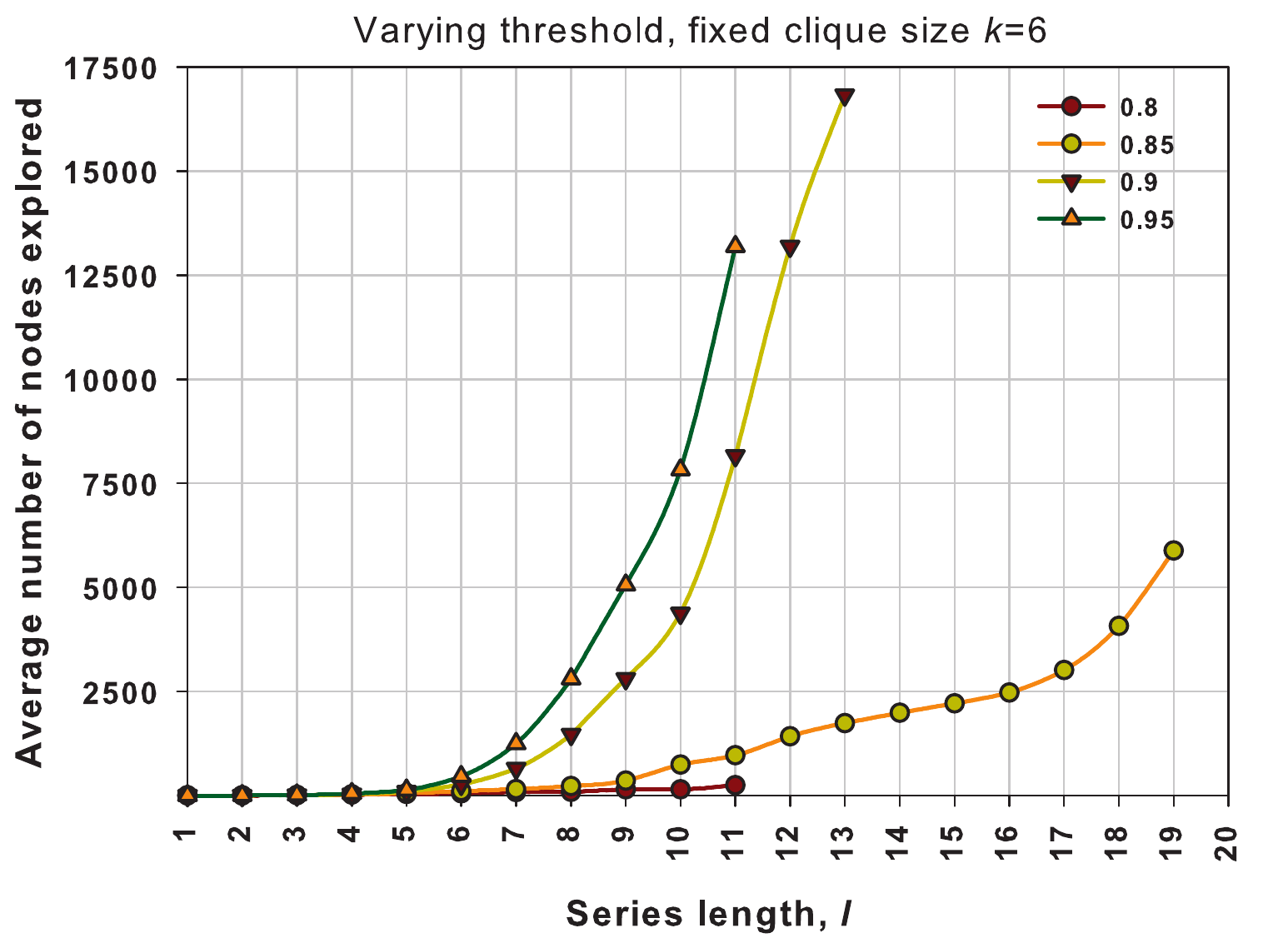}
\includegraphics[scale=0.38]{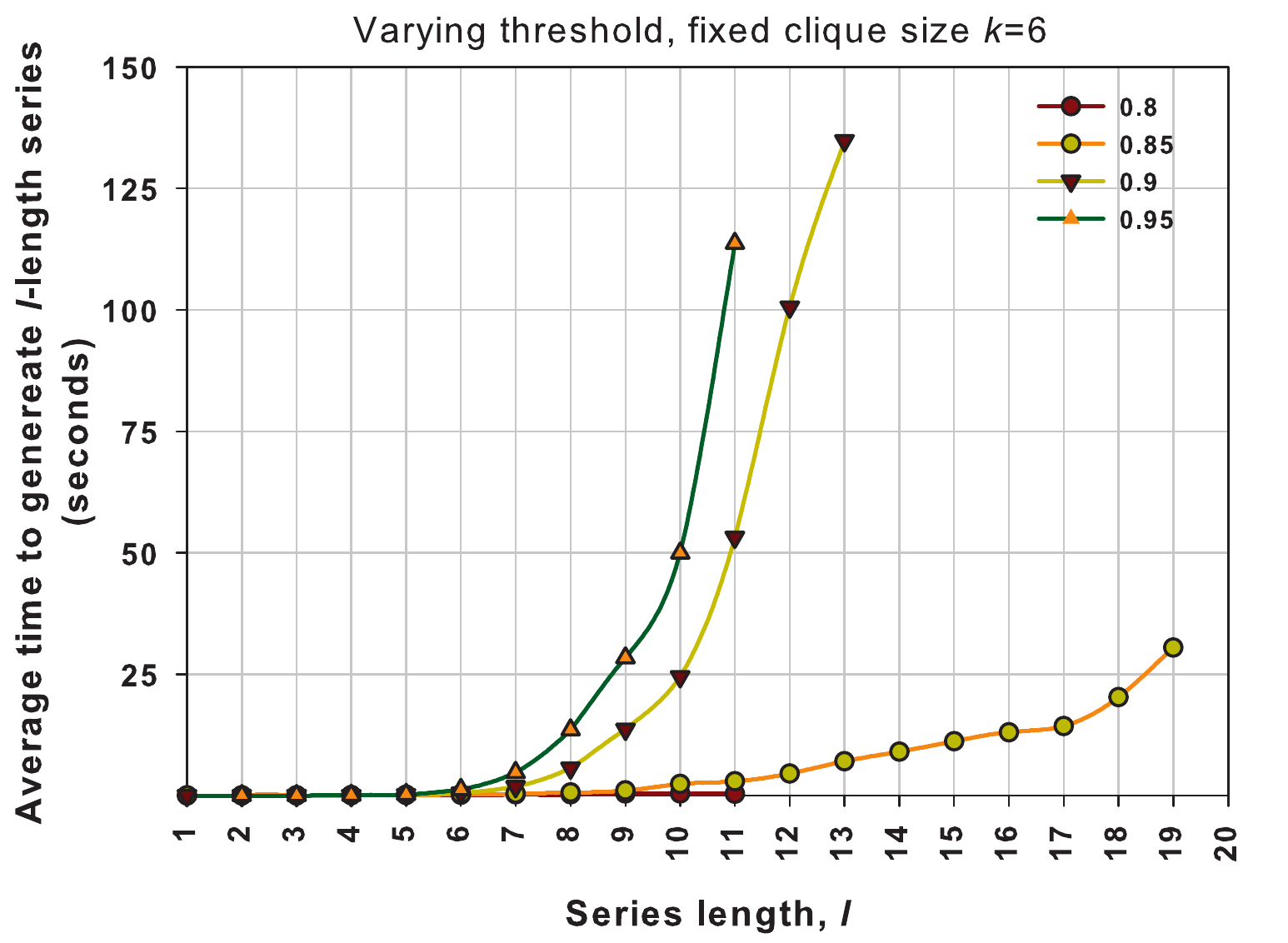} \includegraphics[scale=0.38]{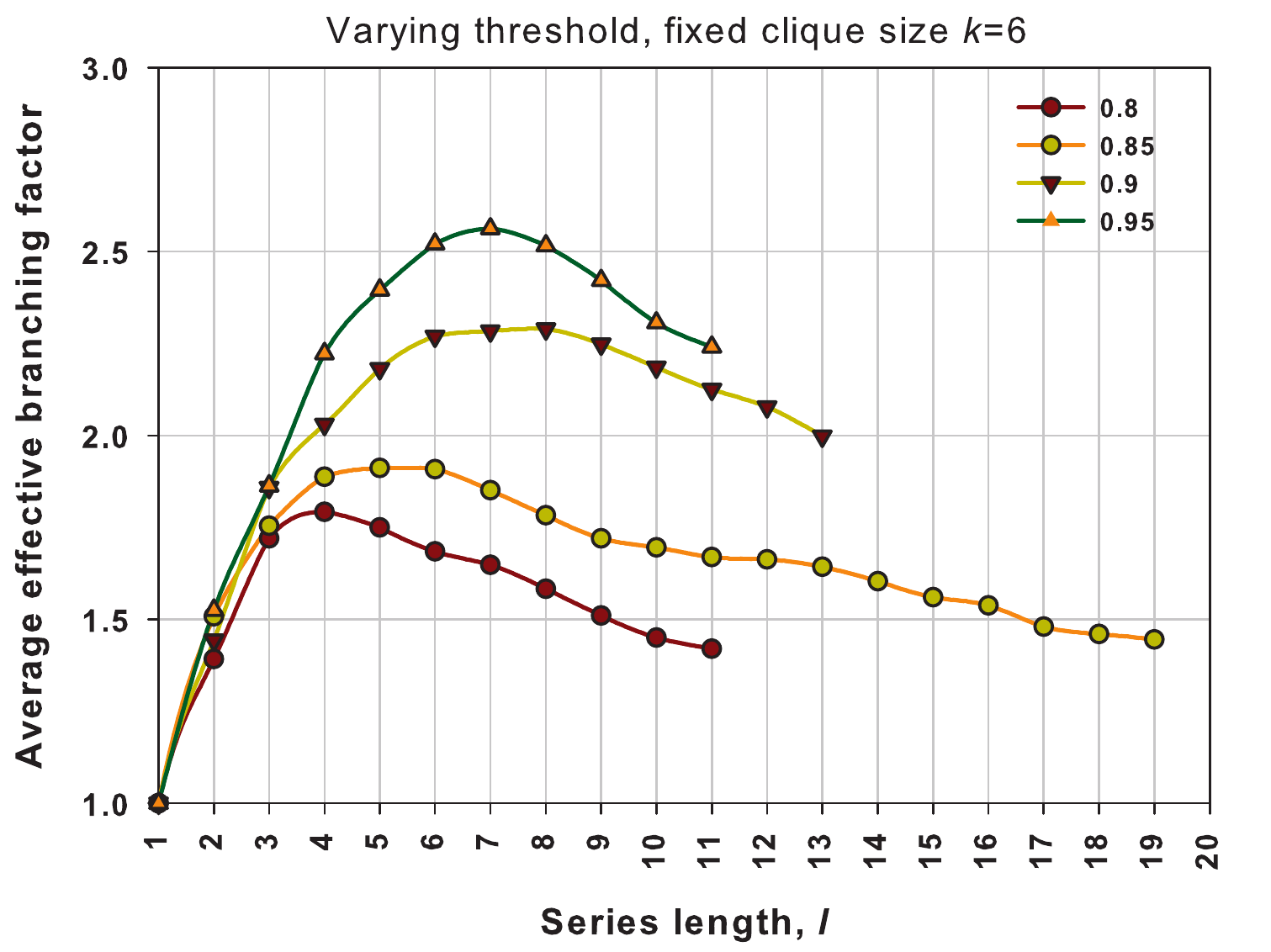}
}

\subfigure[The use of Soergel distance heuristic results in better performance in terms of number of nodes explored, runtime and effective branching factor. The dashed line in each of the plots shows the percentage improvement due to the use of heuristic over plain BFS with $h$=0.]{
\label{fig:PubMed_HeuristicTest} \centering \includegraphics[scale=0.35]{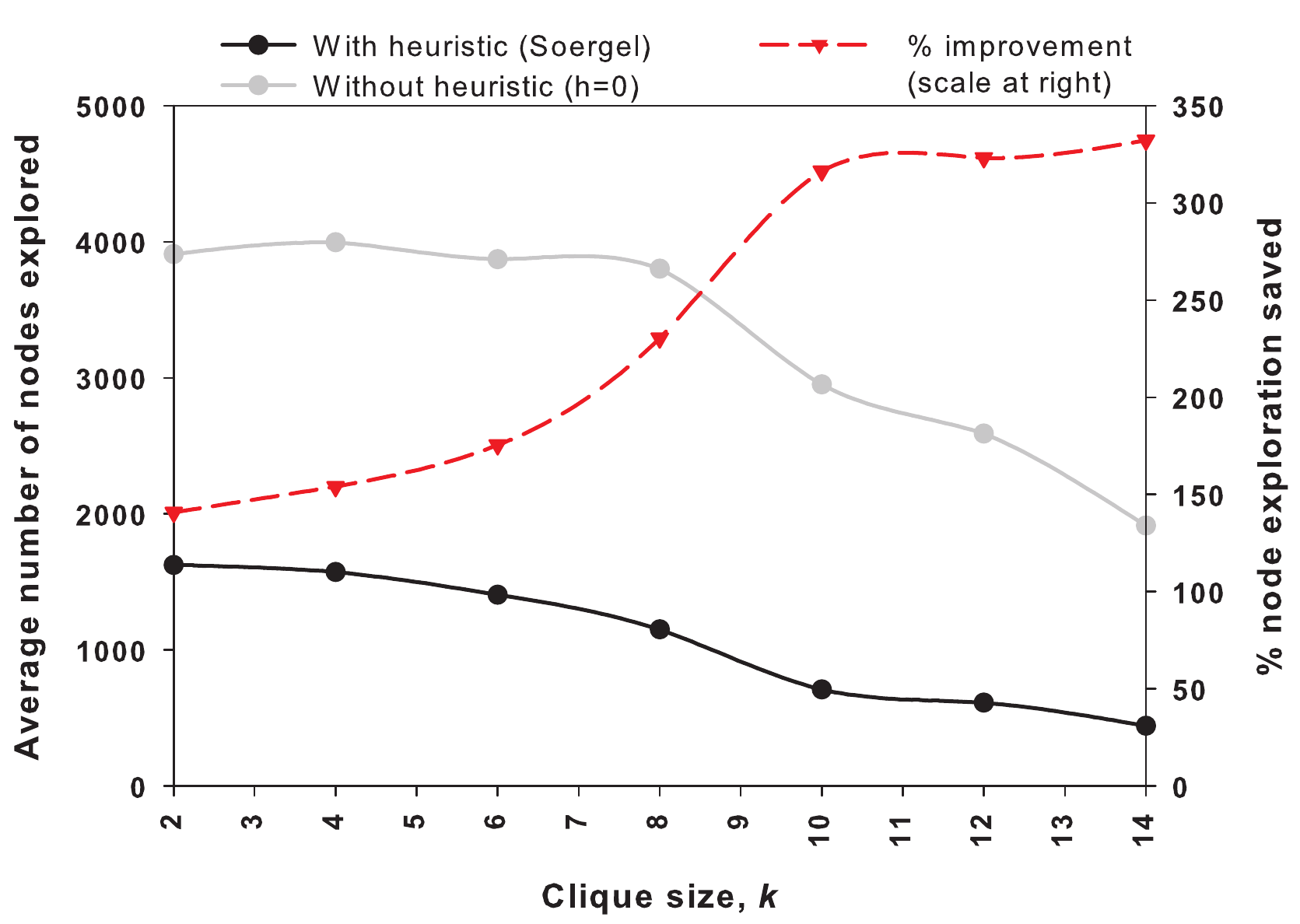}
\includegraphics[scale=0.35]{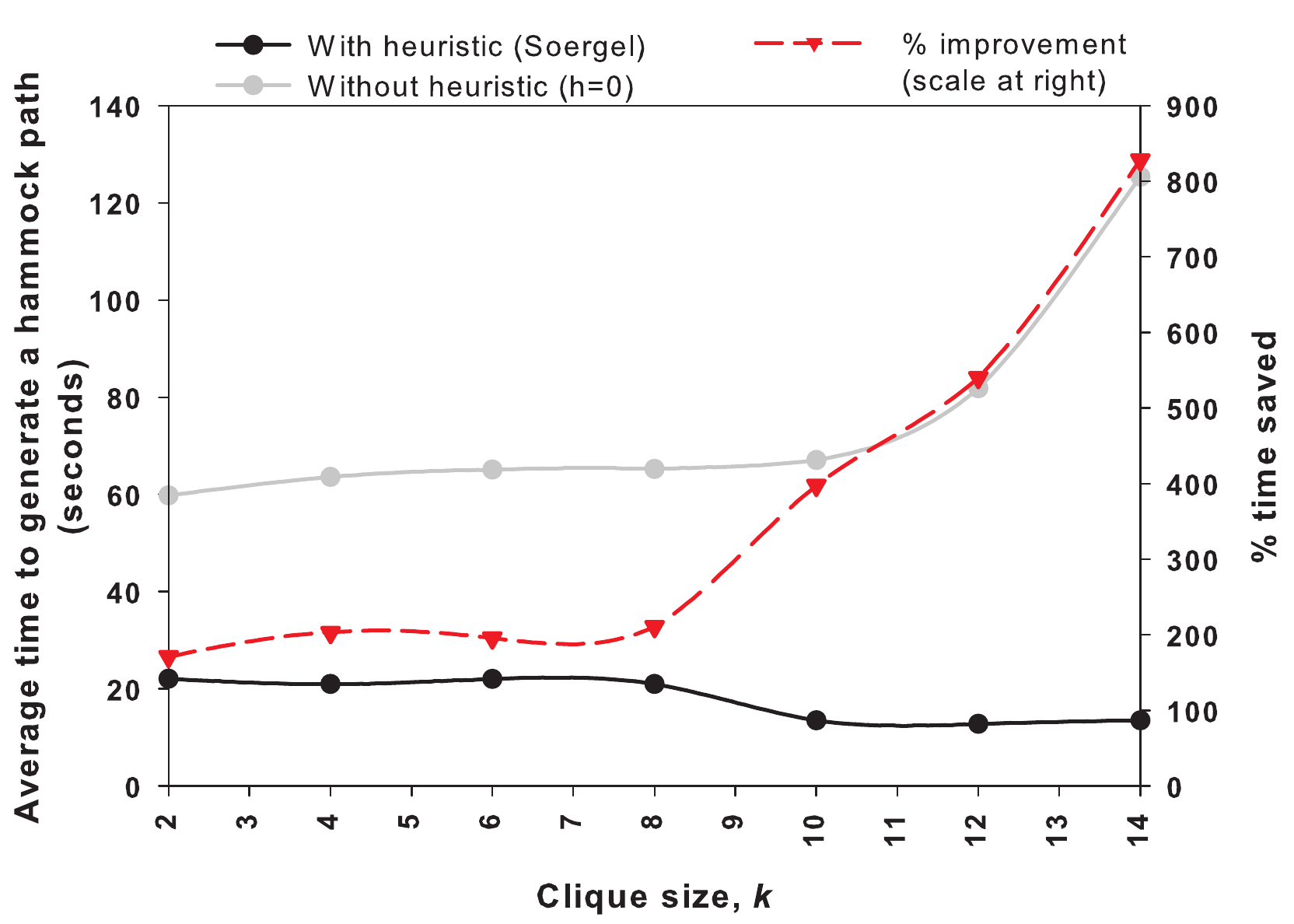}
\includegraphics[scale=0.35]{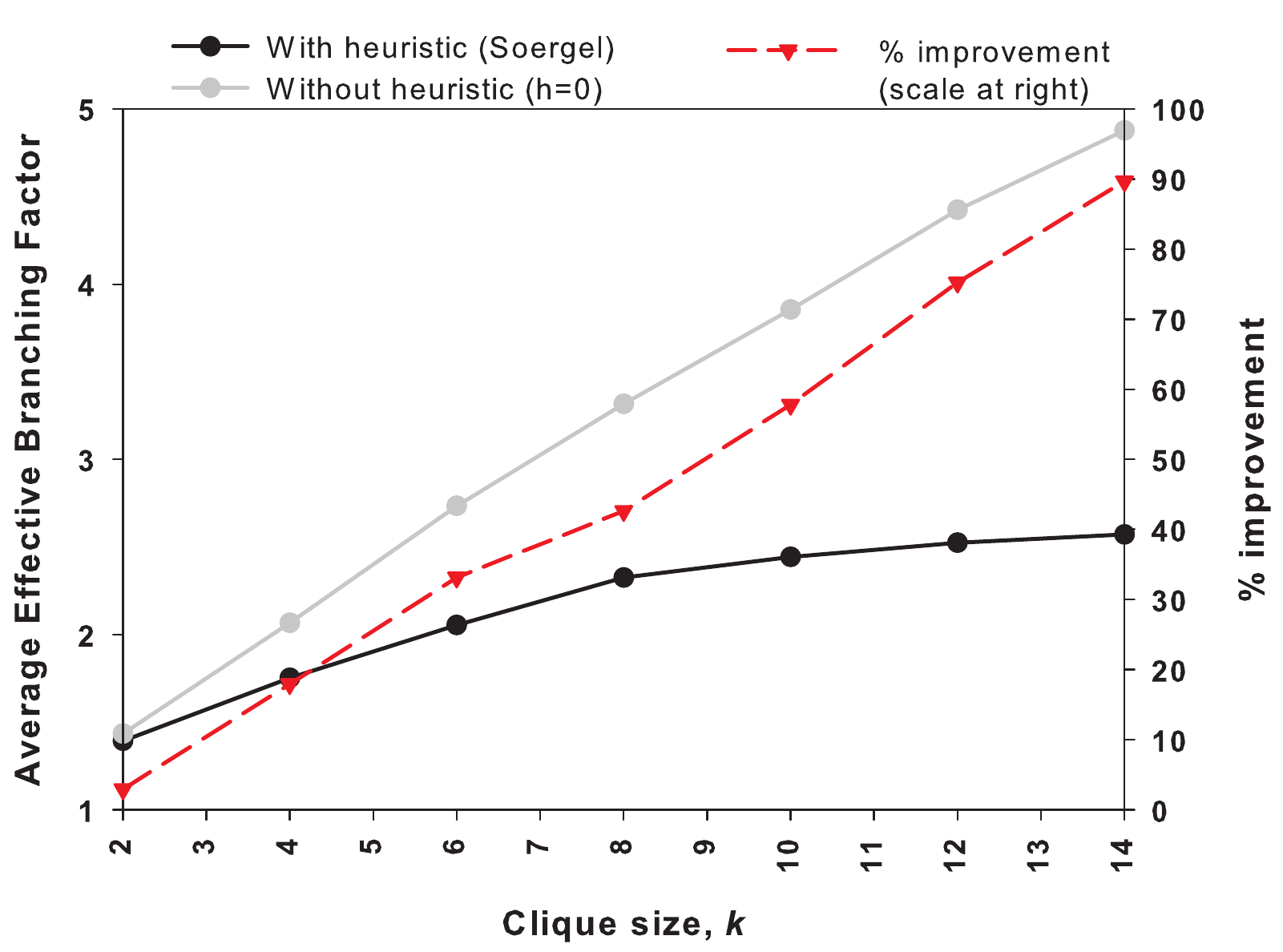}
}\vspace{-0.1in}

\subfigure[Mixed mode operation: (Left) the mixed mode heuristic provides lesser traversal in the similarity network over the corresponding BSF. (Middle) the mixed mode heuristic saves larger percentage of time with larger cliques over the corresponding BFS. (Right) The effective branching factor improvement is higher with larger cliques.]{
\label{fig:Mixed-mode_Pubmed} \centering \includegraphics[scale=0.35]{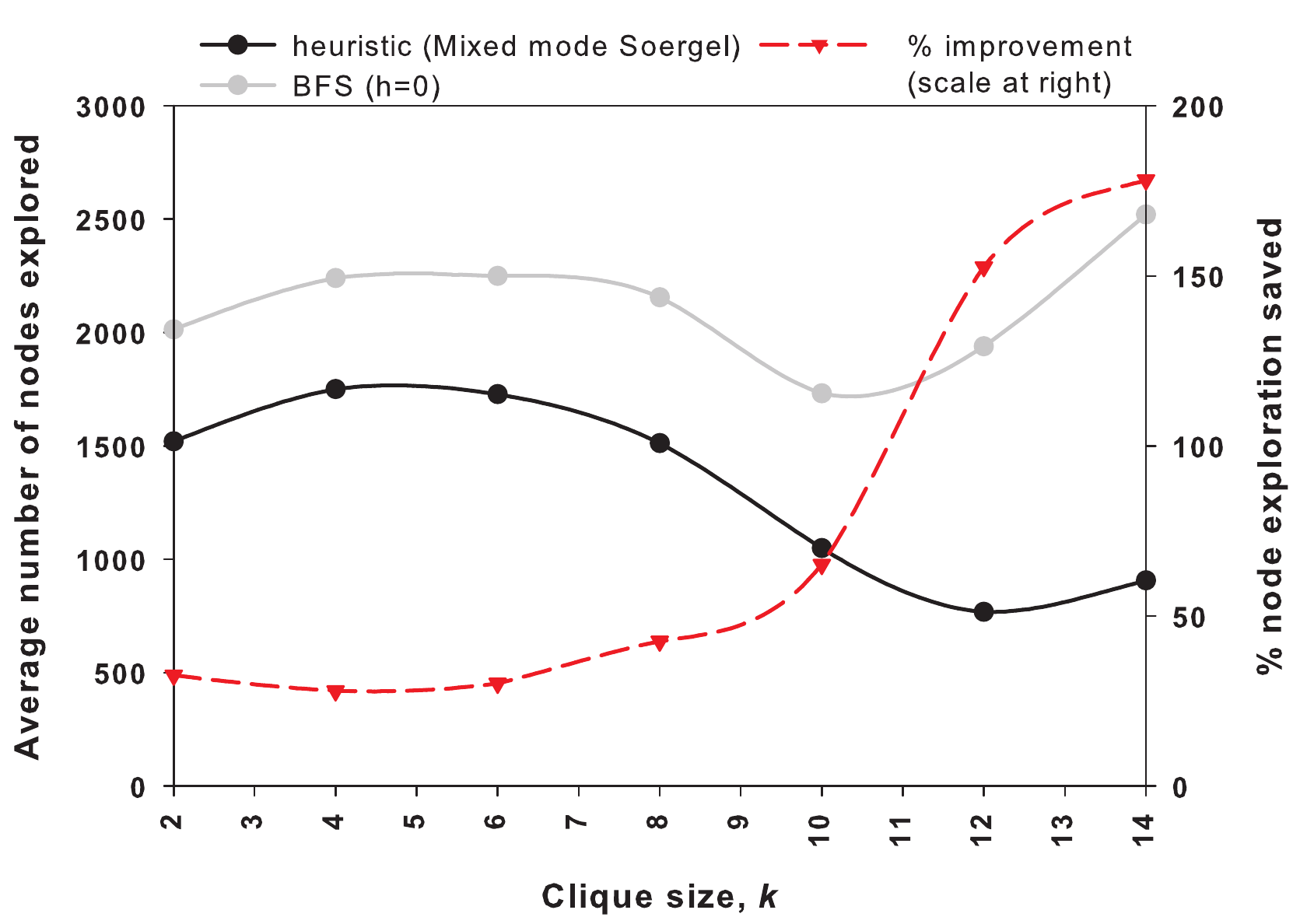}
\includegraphics[scale=0.35]{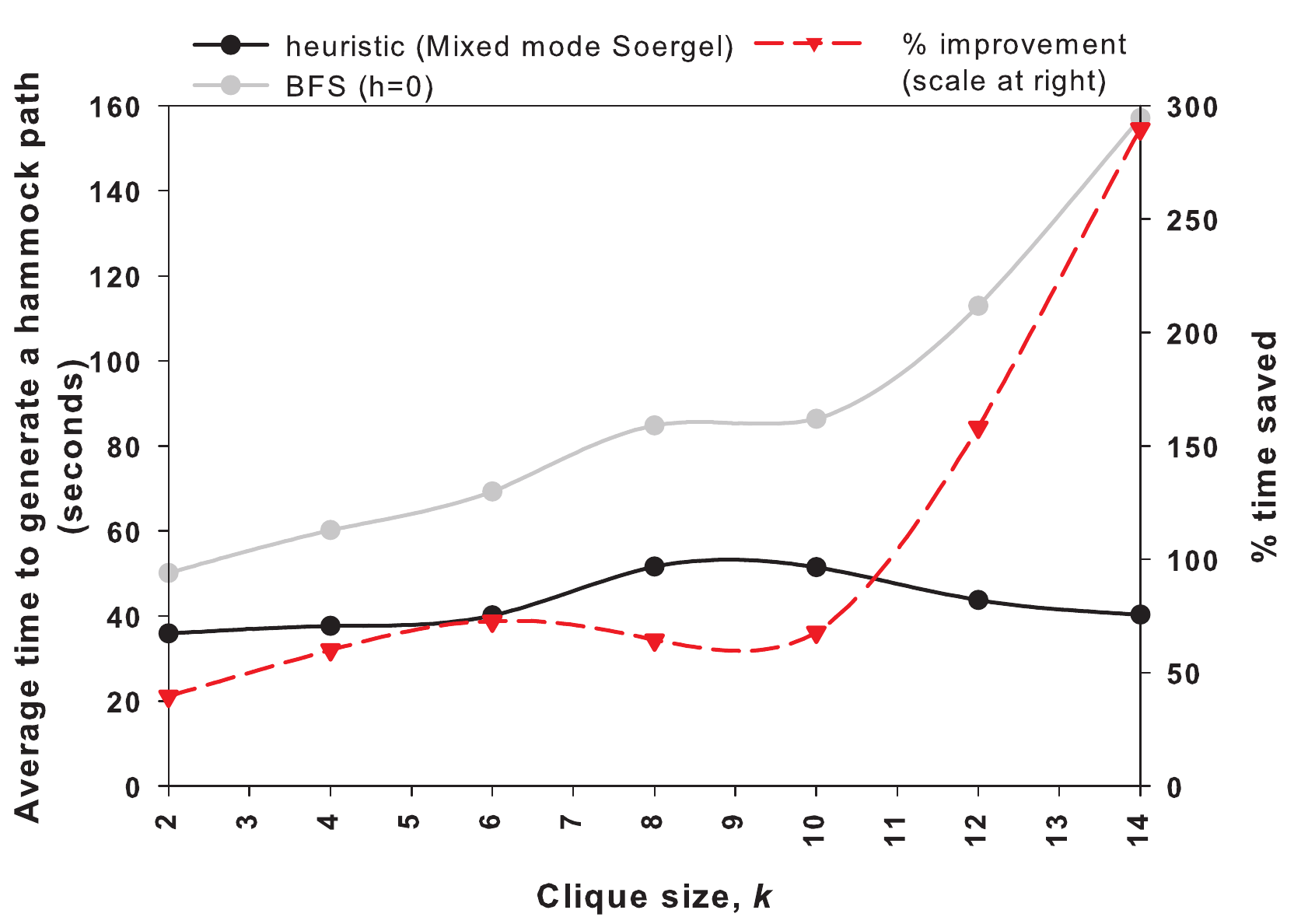}
\includegraphics[scale=0.35]{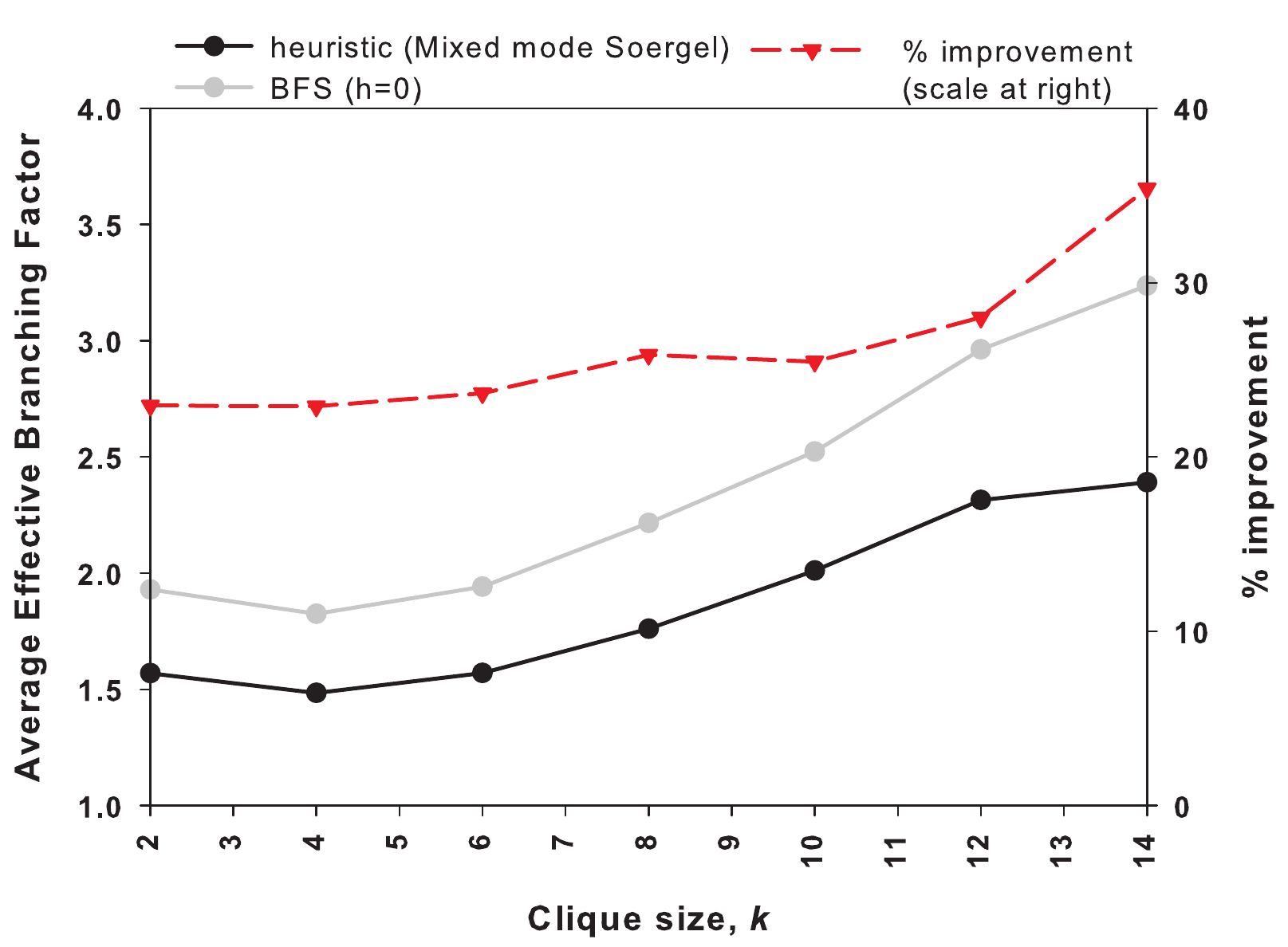}
}\vspace{-0.1in}

\caption{\small{Some illustrative experimental results using the PubMed dataset.}}


\end{figure*}

In our PubMed case study, we view paper abstracts as objects and terms
as features. We curated 161,693 publications related to several cytokines,
transcription factors and feedback molecules and modeled only their
titles and abstracts. There were around 133,252 unique stemmed terms
in the dataset after the removal of stop words, numerals, DNA sequences,
and special characters. Again, for randomly selected pairs of papers,
we discovered over a million hammock paths.

Figure~\ref{fig:Varying-threshold-fixedK_PubMed} shows the trends
of number of nodes explored, time to generate paths, and effective
branching factor, as a function of path length. 
Although Figure \ref{fig:Varying-threshold-fixedK_PubMed} shows that
the average number of nodes explored is large with higher threshold,
it is not necessary that this behavior would remain the same for other
datasets. The result can in fact be opposite in some datasets where
number of paths generated are not lessened by reducing $\theta$,
especially when each object has a large number of features.

Figure \ref{fig:PubMed_HeuristicTest} (left) shows that the use of
Soergel distance heuristic saves significant object exploration over
the vanilla BFS. The larger the clique size, the higher the percentage
by which the heuristic reduces the number of node explorations. It
shows that the use of the straight line Soergel distance as the heuristic
saves more than 300\% node exploration by the A{*} procedure over
the BFS (h=0), for a clique size $k$=14. The saved amount is more
than 100\% even with the smallest clique size $k$=2. The average
time to generate hammock path also reduces due to the saved node exploration.
Figure~\ref{fig:PubMed_HeuristicTest} (middle) shows that the heuristic
saves more than 800\% runtime with clique size $k$=14. Even with
a clique size $k$=2, the savings are near 200\%. In the best case
with $k$=14, the heuristic also improves the effective branching
factor by 90\% over the BFS shown in the plot at the right. In the
worst case with $k$=2, it offers around 4\% improvement of the effective
branching factor.

Despite the use of truncated dataset, Figure \ref{fig:Mixed-mode_Pubmed}
shows that the mixed mode gains due to the heuristic over the BFS
have a similar trend to the normal mode of Figure \ref{fig:PubMed_HeuristicTest}.
Therefore, the mixed mode offers a practical mechanism to provide
the best possible gains from lossy datasets without time consuming
remodeling of the vector space (e.g., \cite{Kumar_Algorithms} uses
costly remodelling as a post processing step). \comment{ \narenc{What
does remodeling mean?}}


\SubSection{Clinical Trials}


\comment{%
\begin{figure}
\centering\includegraphics[scale=0.12]{figs/kidneyheart}

\caption{A clique chain ($k$=5) between two disparate studies on patients
with Congestive Heart Failure and patients with kidney problems. \label{fig:hammock-chain_kidney_Heart}}

\end{figure}

}

We curated more than 60 thousand clinical trials from clinicaltrials.gov
and concentrated on the purpose and description of each trial. Here
trials are objects and terms are features. The concept lattice we
used had around a thousand unique concepts and was generated with
minsup=10\%. Due to space restrictions, we show qualitative rather
than quantitative results here. 

\comment{Figure \ref{fig:hammock-chain_kidney_Heart} shows a clique
chain with clique size $k$=5. The labels in the junction nodes signify
the study-goal of the trials. The detailed series composed of start
object, junction points of the cliques and the goal object is given
in Table \ref{tab:KidneyHeart}. }

\begin{table}[!t]

\caption{\small{Clinical trials: A hammock path connecting 
congestive heart failure and kidney complications.} \label{tab:KidneyHeart}}

\centering \begin{tabular}{|>{\centering}p{0.5in}|>{\raggedright}p{5.0in}|}
\hline 
\textbf{\small Trial ID}{\small {}}  & \textbf{\small Short Title of the Trial}\tabularnewline
\hline 
{\small 00103519}  & {\small Study of DITPA in Patients With Congestive Heart Failure.}\tabularnewline
\hline 
{\small 00696631}  & {\small European Trial of Dronedarone in Moderate to Severe Congestive
Heart Failure.}\tabularnewline
\hline 
{\small 00697086}  & {\small \comment{European} Study of Dronedarone in Atrial Fibrillation.}\tabularnewline
\hline 
{\small 00744874}  & {\small Ablation of the Pulmonary Veins for Paroxysmal Afib.}\tabularnewline
\hline 
{\small 00807586}  & {\small Corticosteroid Pulse After Ablation.}\tabularnewline
\hline 
{\small 00030563}  & {\small Surgery With or Without Radio frequency Ablation Followed
by Irinotecan in Treating Patients With Colorectal Cancer that is
Metastatic to the Liver.}\tabularnewline
\hline 
{\small 00003753}  & {\small Floxuridine, Dexamethasone, and Irinotecan After Surgery in
Treating Patients With Liver Metastases From Colorectal Cancer.}\tabularnewline
\hline 
{\small 00005818}  & {\small SU5416 and Irinotecan in Treating Patients With Advanced Colorectal
Cancer.}\tabularnewline
\hline 
{\small 00002828}  & {\small Chemotherapy With Raltitrexed and Fluorouracil in Treating
Patients With Advanced Colorectal Cancer.}\tabularnewline
\hline 
{\small 00449137}  & {\small Arsenic Trioxide, Fluorouracil, and Leucovorin in Treating
Patients With Stage IV Colorectal Cancer That Has Relapsed or Not
Responded to Treatment.}\tabularnewline
\hline 
{\small 00124605}  & {\small Arsenic Trioxide and Pamidronate in Treating Patients With
Advanced Solid Tumors or Multiple Myeloma.}\tabularnewline
\hline 
{\small 00302627}  & {\small Pamidronate, Vitamin D, and Calcium for the Bone Disease of
Kidney and Heart Transplantation.}\tabularnewline
\hline 
{\small 00074516}  & {\small Kidney Transplantation in Patients With Cystinosis.}\tabularnewline
\hline
\end{tabular}


\end{table}

Table \ref{tab:KidneyHeart} describes a significant chain of trials
connecting two disparate clinical studies related to congestive heart
failure and kidney transplantation in patients with cystinosis. 
(One accepted practice to assess the
statistical significance of discovered chains is, for each step
in the path, to assess the likelihood of overlap for the given
descriptor sizes using the hypergeometric distribution and attribute
a $p$-value after FDR corrections such the Benjamini-Hochberg procedure.)
The chain starts with heart failure trials and goes through studies on
atrial fibrillation (abnormal heart rhythm), cardiac ablation, colorectal
cancer, advanced solid tumors, and eventually reaches clinical study
on kidney transplantation in patients with cystinosis. Such connections
between cardio-vascular disease and kidney failures are an intense topic
of current research (e.g., see~\cite{Gill_Outcomes}).


\vspace{-0.1in}
 \Section{Discussion} \vspace{-0.15in}
 We have presented an efficient algorithmic approach to discover hammock
paths in similarity networks without inducing these networks in their
entirety. The experimental results have demonstrated scalability,
effectiveness of our heuristics, and ability to yield domain-specific
insight. We posit that our approach can be a useful information exploration
tool for understanding the structure of connectivities underlying
boolean and vector-valued association datasets. Future work is geared
toward incorporating additional distance measures and defining new
compressed representations of datasets that can serve multiple uses,
from concept modeling to distance estimation. 


\end{document}